\documentclass[sw, crcready]{iosart2x}

\usepackage{dcolumn}

\usepackage[numbers]{natbib}
\usepackage{hyperref}
\hypersetup{
  hidelinks,
  colorlinks=true,
  allcolors=black
}
\usepackage{caption}
\usepackage{mathtools}
\usepackage{booktabs}
\usepackage{multirow}
\usepackage{multicol}
\usepackage{color}
\usepackage{float}
\usepackage{blkarray}
\usepackage{graphicx}
\usepackage{subfig}
\graphicspath{{./}}

\newcommand{\aka}{\textit{a}.\textit{k}.\textit{a}.\ }
\newcommand{\ie}{\textit{i}.\textit{e}.\ }
\newcommand{\eg}{\textit{e}.\textit{g}.\ }

\def\highlightcolor{black}
\newcommand{\marktext}[1]{\textcolor{\highlightcolor}{#1}}
\newcommand{\blacktext}[1]{\textcolor{black}{#1}}

\DeclareCaptionFormat{customCaption}{
\footnotesize#1 #3
}

\def\tsc#1{\csdef{#1}{\textsc{\lowercase{#1}}\xspace}}
\tsc{WGM}
\tsc{QE}

\newcolumntype{d}[1]{D{.}{.}{#1}}

\firstpage{1}
\lastpage{4}
\volume{1}
\pubyear{0000}

\begin{document}
\begin{frontmatter} 

\newtheorem{definition}{Definition}

\title{DegreEmbed: incorporating entity embedding into logic rule learning for knowledge graph reasoning}

\runtitle{DegreEmbed: incorporating entity embedding into logic rule learning for knowledge graph reasoning}

\begin{aug}
\author[A]{\inits{H.}\fnms{Haotian} \snm{Li}\ead[label=e1]{lcyxlihaotian@126.com}}
\author[A]{\inits{H.}\fnms{Hongri} \snm{Liu}\ead[label=e2]}
\author[A]{\inits{Y.}\fnms{Yao} \snm{Wang}\ead[label=e3]}
\author[A]{\inits{G.}\fnms{Guodong} \snm{Xin}\ead[label=e4]}
\author[A]{
  \inits{Y.}\fnms{Yuliang} \snm{Wei}\ead[label=e5]{wei.yl@hit.edu.cn}
  \thanks{Corresponding author.}
}
\address[A]{
  School of Computer Science and Technology, \orgname{Harbin Institute of Technology at Weihai},
  \cny{China}
  \printead[presep={\\}]{e1,e5}
}
\end{aug}


\begin{abstract}
Knowledge graphs (KGs), as structured representations of real world facts, are intelligent databases
incorporating human knowledge that can help machine imitate the way of human problem solving.
However, KGs are usually huge and there are inevitably missing facts in KGs, \blacktext{thus undermining
applications such as question answering and recommender systems that are based on knowledge graph reasoning.}
Link prediction for knowledge graphs is the task aiming to complete missing facts by reasoning based on
the existing knowledge. Two main streams of research are widely studied: one learns low-dimensional
embeddings for entities and relations that can explore latent patterns, and the other gains good interpretability
by mining logical rules.
\marktext{
Unfortunately, the heterogeneity of modern KGs that involve entities and relations of various types is not well considered in the previous studies.
}
In this paper, we propose DegreEmbed, a model that combines embedding-based learning and logic rule
mining for inferring on KGs. Specifically, we study the problem of predicting missing links in
heterogeneous KGs from the perspective of
the degree of nodes. Experimentally, we demonstrate that our DegreEmbed model outperforms
the state-of-the-art methods on real world datasets and the rules mined by our model are
of high quality and interpretability.
\end{abstract}


\end{frontmatter}

\begin{multicols}{2}

\section{Introduction}
Recent years have witnessed the growing attraction of knowledge graphs in a variety of applications,
such as dialogue systems \cite{liu2019knowledge, moon2019opendialkg}, search engines \cite{xiong2017explicit}
and domain-specific softwares \cite{sousa2020evolving, mohamed2020discovering}. Capable of incorporating
large-scale human knowledge, KGs provide graph-structured representation of data that can be
comprehended and examined by humans. Knowledge in KGs is stored in triple form $(e_s, r, e_o)$,
with $e_s$ and $e_o$ denoting subject and object entities and $r$ a binary relation (\aka predicate).
For example, the fact that Mike is the nephew of Pete can be formed as
$(\texttt{Mike}, \texttt{nephewOf}, \texttt{Pete})$. However, \blacktext{information incompleteness can
be seen in most modern KGs}, that is, missing links in the graph, \textit{e.g.},
the work of \cite{dong2014knowledge, krompass2015type} shows that there are more than 66\%
of the person entities missing a birthplace in two open KGs
Freebase \cite{bollacker2008freebase} and DBpedia \cite{auer2007dbpedia}.

\begin{figure*}[t]
  \centering
  \includegraphics[scale=.4]{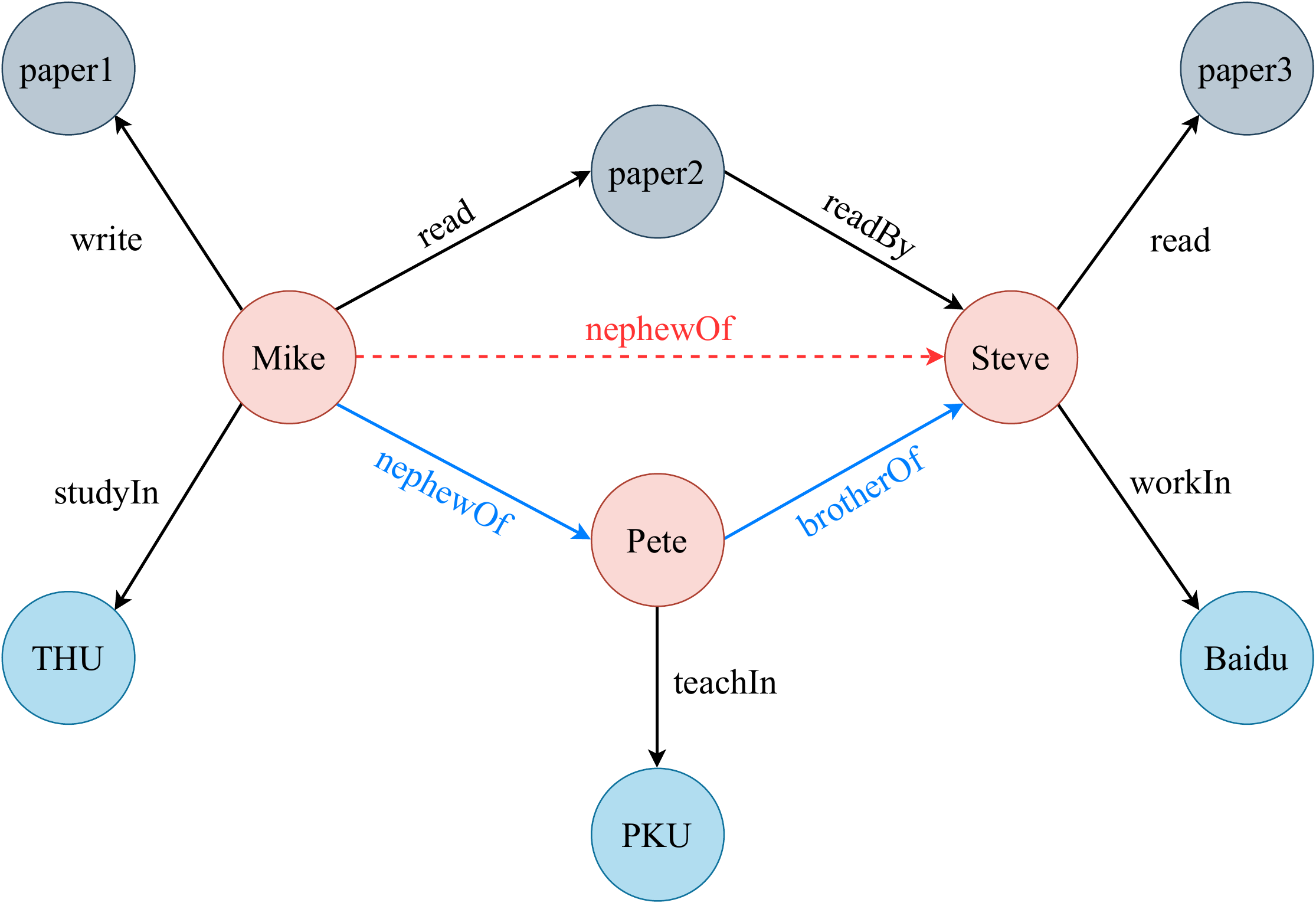}
  \caption{
    An example of KG containing heterogeneous entities and relations: paper, person and institution.
    Entities in different colors mark their type. The existing links in the KG are presented
    as solid black lines, the missing one as dashed lines in red and the proper rule for
    inferring the link as blue lines.
  }
  \label{fig:heterogeneous}
\end{figure*}

Predicting missing triples based on the existing facts is usually called \emph{link prediction}
as a subtask of knowledge graph completion (KGC) \cite{ji2021survey}, and numerous models
have been developed for solving such problems. One prominent direction in this line of research
is \emph{representation learning} methods that learn distributed embedding vectors for
entities and relations, such as TransE \cite{bordes2013translating} and ComplEx \cite{trouillon2016complex}.
In this work, they are referred to as embedding-based methods.
This kind of models are capable of capturing latent knowledge through low-dimensional vectors,
\textit{e.g.} we can classify male and female entities in a family KG by clustering their
points at the semantic space. In spite of achieving high performance, these models suffer from
non-transparency and can poorly be understood by humans, which is a common issue for most
deep learning models. \blacktext{In addition, most embedding-based methods work in a transductive setting,
where they require the entities in training and test data to overlap, hindering the way to generalize
in some real-world situations.}

Another popular approach is rule mining that discovers logical rules through mining co-occurrences
of frequent patterns in KGs \cite{chen2016ontological, galarraga2015fast}. This paper studies
the problem of learning first-order logical Horn clauses for knowledge graph reasoning (KGR).
As illustrated in Fig. \ref{fig:heterogeneous}, there is a missing link (\ie \texttt{nephewOf})
between the subject \texttt{Mike} and the object \texttt{Steve}, but we can complete the fact
through a logic rule \texttt{nephewOf}(\texttt{Mike}, \texttt{Pete}) $\wedge$
\texttt{brotherOf}(\texttt{Pete}, \texttt{Steve}) $\Rightarrow$
\texttt{nephewOf}(\texttt{Mike}, \texttt{Steve}), meaning that if Mike is the nephew
of Pete and Steve has a brother Pete, then we can infer that Mike is the nephew of Steve.
Reasoning on KGs through Horn clauses has been previously studied in the area of
\emph{Inductive Logic Programming} \cite{muggleton1994inductive}. One representative method,
Neural LP \cite{yang2017differentiable}, is the first fully differentiable neural system that
successfully combines learning discrete rule structures as well as confidence scores in
continuous space. Although learning logical rules equips a model with strong interpretability
and the ability to generalize to similar tasks \cite{qu2019probabilistic, zhang2020efficient},
these methods often focus only on the relations of which the rules are made up, while the
intrinsic properties of the involved entities are not considered. For example, in the KG
shown in Fig. \ref{fig:heterogeneous}, it is definitely wrong to infer by a rule containing
a female-type relation path like \texttt{sisterOf} starting from Mike, because Mike is the nephew of Pete,
which indirectly tells us he is a male. This sort of deficiency is more severe in heterogeneous
KGs where there are entities and relations of different types mixing up. In these KGs,
\marktext{
there might be multiple rules of no use, becoming inevitable noises for reasoning tasks. For instance, the rule \texttt{read}(\texttt{Mike}, \texttt{paper2}) $\wedge$ \texttt{readBy}(\texttt{paper2}, \texttt{Steve}) $\Rightarrow$ \texttt{nephewOf}(\texttt{Mike}, \texttt{Steve}) is obviously wrong in logic, which might decrease the performance and interpretability of ILP models.
}

In this paper, in order to bridge the gap between the two lines of research mentioned above,
we propose DegreEmbed, a model of logic rule learning that integrates the inner attributes of
entities by embedding nodes in the graph from the perspective of their degrees. DegreEmbed is not
only interpretable to humans, but also able to mine relational properties of entities.
We also evaluate our model on several knowledge graph datasets, and show that we are able to learn
more accurately, and meanwhile, gain strong interpretability of mined rules.

\blacktext{
Our main contributions are summarized below:
}
\begin{itemize}
  \item We propose an original model based on logic rule learning to predict missing links in heterogeneous KGs.
        Specifically, a new technique for encoding entities, called degree embedding, is designed to capture
        hidden features through the relation type of edges incident to a node.
  \item Comparative experiments on knowledge graph completion task with five benchmark datasets prove
        that our DegreEmbed model outperforms baseline models. Besides, under the evaluation of a metric
        called Saturation, we show that our method is capable of mining meaningful logic rules from 
        knowledge graphs.
  \item Visualizing learned entity embeddings, we demonstrate that clear features of entities can be obtained
        by our model, thus benefitting the prediction in heterogeneous settings. Moreover, we prove the necessity
        of each component of our model using ablation study.
\end{itemize}

\blacktext{
This paper is structured as follows. We briefly introduce our related work and review
preliminary definitions of knowledge graphs respectively in Section \ref{sec-related}
and Section \ref{sec-preliminaries}. Section \ref{sec-method} introduces our proposed DegreEmbed model
based on logic rule learning for link prediction in heterogeneous KGs. We present the experimental results
in Section \ref{sec-experiment} and conclude our work by pointing out the future direction.
}

\section{Related work} \label{sec-related}

\blacktext{
Our work is first related to previous efforts on relational data mining, based on which, a large body of
deep rule induction models have been developed for link prediction. Since our approach achieves a combination
of logic rule learning and knowledge graph embedding, we conclude related work in this topic as well.
}

\textbf{Relational data mining}. The problem of learning relational rules has been traditionally
addressed in the field of \emph{inductive logic programming} (ILP) \cite{muggleton1994inductive}.
These methods often view the task of completing a missing triple as a query $q(h, t)$ where
they learn a probability as confidence score for each rule between the query entity and answer entity.
Among these studies, Path-Ranking Algorithm (PRA) \cite{lao2010relational} investigated the framework
of random walk inference, where they select a relational path under a set of constraints and
perform maximum-likelihood classification. An RNN model was developed by \cite{neelakantan2015compositional}
to compose the semantics of relations for arbitrary-length reasoning. Chain-Reasoning proposed by
\cite{das2016chains}, enabling multi-hop reasoning through a neural attention mechanism,
reveals logical rules across all relations and entities. Although ILP models are capable of
mining interpretable rules for humans, these models typically take both positive and negative
examples for training and suffer from a potentially large version space, which is a critical
shortage since most modern KGs are huge and contain only positive instances.

\textbf{Neural logic programming}. \blacktext{In recent years, models borrowing the idea of logic rule learning
in a deep manner have emerged as successful approaches for link prediction task.}
Extending the idea of TensorLog that tackles the problem of rule-based logic reasoning through
sparse matrix multiplication, Neural LP \cite{yang2017differentiable} is the first end-to-end
differentiable approach to simultaneously learn continuous parameters and discrete structure of rules.
Some recent methods \cite{sadeghian2019drum, wang2019differentiable, yang2019learn}
have improved the framework done by Neural LP \cite{yang2017differentiable} in different manners.
DRUM \cite{sadeghian2019drum} introduces tensor approximation for optimization and reformulate
Neural LP to support rules of varied lengths. Neural-Num-LP \cite{wang2019differentiable}
extends Neural LP to learn numerical relations like \emph{age} and \emph{weight} with dynamic
programming and cumulative sum operation. NLIL \cite{yang2019learn} proposes a multi-hop reasoning
framework for general ILP problem through a divide-and-conquer strategy as well as
decomposing the search space into three subspaces. However, the existing methods ignore the effects
\marktext{caused by entities of different types} while reasoning over a specific relational path, thus witness a more
obvious failure where heterogeneous entities and relations are involved in the KGs.

\textbf{Representation learning}. Capturing their semantic information by learning low-dimensional
embeddings of entities and relations, also known as \emph{knowledge graph embedding},
is a vital research issue in KGC, and we term those models as embedding-based models.
Newly proposed methods, including RotatE \cite{sun2019rotate}, ConvE \cite{dettmers2018convolutional}
and TuckER \cite{balavzevic2019tucker}, predict missing links by learning embedding vectors from
various perspectives of the problem. Specifically, the work of RotatE \cite{sun2019rotate}
focuses on inferring patterns such as symmetry and inversion, where they proposed a rotational
model that rotates the relation from the subject to the object in the complex space as
$e_o = e_s \circ r$ where the $\circ$ denotes the element-wise Hadamard product.
ConvE introduces a highly parameter efficient model, which uses 2D convolution over embeddings and
multiple layers of nonlinear features to express semantic information.
TuckER, inspired by the Tucker decomposition \cite{tucker1966some} that factorizes a tensor
into a core tensor along with a set of matrices, is a linear model for link prediction that has
good expressive power. Unfortunately, the biggest problem is that these sort of methods can
hardly be comprehended by humans, but we relate to these methods for their ability to capture
latent information of entities and relations through embedding.

We also notice that there are methods trying to establish a connection between learning logical rules
and learning embedding vectors \cite{zhang2019iteratively, wangLogicRulesPowered2019}, 
\marktext{
where they augment the dataset by
exploring new triplets from the existing ones in the KG using pre-defined logical rules to deal with the sparsity problem, which differs from our goal to consider entities of various types for learning in heterogeneous KGs.
}

\section{Preliminaries} \label{sec-preliminaries}

\blacktext{
In this section, we introduce background concepts of logic rule learning for knowledge graph
reasoning as well as the definition of topological structure of KGs.
}

\subsection{Knowledge graph reasoning}
\textbf{Knowledge graph} can be modeled as a collection of factual triples
$\mathcal{G} = \{ (e_s, r, e_o)\ |\ e_s, e_o \in \mathcal{E}, r \in \mathcal{R} \}$,
with $\mathcal{E}, \mathcal{R}$ representing the set of entities and binary relations
(\aka predicates) respectively in the knowledge graph, and $(e_s, r, e_o)$ the triple
$(\texttt{subject}, \texttt{relation}, \texttt{object})$ in form of $e_s \mathop{\rightarrow}\limits^r e_o$.
During reasoning over KGs, each triple is usually presented in the form $\texttt{r}(e_s, e_o)$.
The subgraph regarding a specific relation $\texttt{r}_i$ is described as a subset of
$\mathcal{G}$ containing all triples with $\texttt{r}_i$ being the connection between the subject
and object: $\mathcal{G}(\texttt{r}_i) = \{ (e_s, r, e_o)\ |\ e_s, e_o \in \mathcal{E},
\texttt{r}_i \in \mathcal{R}, r = \texttt{r}_i\}$.

\textbf{Logic rule reasoning}. We perform reasoning on KGs by learning a \emph{confidence} score
$\alpha \in [0,1]$ for a first-order logic \emph{rule} in the form
\begin{equation}
\texttt{r}_1(x, z_1) \wedge \cdots \wedge \texttt{r}_l(z_{l-1}, y) \Rightarrow q(x, y)\ : \ \alpha,
\end{equation}
$\textbf{r(x, y)} \Rightarrow q(x, y)$ for short, with $r_1, \ldots, r_{l}, q \in \mathcal{R}$,
$z_1, \ldots, z_{l-1} \in \mathcal{E}$, where $\textbf{r} = \wedge_i r_i$, is called a \emph{rule pattern}.
For example, the rule $\texttt{brotherOf}(x, z) \wedge \texttt{fatherOf}(z, y) \Rightarrow \texttt{uncleOf}(x, y)$
intuitively states that if $x$ is the brother of $z$ and $z$ is the father of $y$,
then we can conclude that $x$ is the uncle of $y$.
All rule patterns of length $l$ ($l \geq 2$) can be formally defined as a set of relational tuples
$\mathcal{H}^l = \{ (r_1, r_2, \ldots, r_l)\ |\ r_i \in \mathcal{R}, 1 \leq i \leq l \} = \mathcal{R}^l$,
and the set of patterns no longer than $L$ is denoted as
$\mathbb{H}^L = \mathop{\cup}\limits_{l=2}^L{\mathcal{H}^l}$.
A \emph{rule path} \textbf{p} is an instance of a certain pattern \textbf{r} via different sequences of
entities, which can be denoted as $\textbf{p} \rhd \textbf{r}$, \textit{e.g.}, $\left(r_a(x, z_1), r_b(z_1, y)\right)$
and $\left(r_a(x, z_2), r_b(z_2, y)\right)$ are different paths of the same pattern.

The link prediction task here is considered to contain a variety of queries,
each of which is composed of a \texttt{query} body $q \in \mathcal{R}$,
an entity \texttt{head} $h$ which the query is about, and an entity \texttt{tail} $t$
that is the answer to the query such that $(h, q, t) \in \mathcal{G}$. Finally we want to find the
most possible logic rules $h \mathop{\rightarrow}\limits^{r_1} \cdots
\mathop{\rightarrow}\limits^{r_l} t$ to predict the link $q$.
Thus, given maximum length $L$, we assign a single \emph{confidence} score (\ie probability)
to a set of rule paths \textbf{p}'s adhering to the same pattern \textbf{r} that connects $h$ and $t$:
\begin{equation}\label{eq:multi-target}
\{ \textbf{p}_i(h, t) \Rightarrow q(h, t)\ |\ \textbf{p}_i \rhd \textbf{r}, \textbf{r} \in
\mathbb{H}^L \} \ : \ \alpha
\end{equation}

During inference, given an entity $h$, the unified score of the answer $t$
can be computed by adding up the confidence scores of all rule paths that infer
$q(h, t)$, and the model will produce a ranked list of entities where higher the
score implies higher the ranking.

\subsection{Graph structure}
\begin{definition}[Directed Labeled Multigraph]
A \textbf{directed labeled multigraph} $G$ is a tuple $G = (V, E)$,
where $V$ denotes the set of vertices, and $E \subseteq V \times V$
is a multiset of directed, labeled vertex pairs (\ie edges) in the graph $G$.
\end{definition}

Because of its graph structure, a knowledge graph can be regarded as a directed labeled multigraph
\cite{stokman1988structuring}. In this paper, "graph" is used to refer to "directed labeled multigraph"
for the sake of simplicity. $G(\texttt{r}) = (V(\texttt{r}), E(\texttt{r}))$
is the corresponding topological structure of $\mathcal{G}(\texttt{r})$.
$m = |V|$ and $n = |E|$ stand for the \textbf{number of vertices}
and \textbf{number of edges} respectively for a graph $G$.
Particularly in a KG, $|\mathcal{E}| = m$ and the total number of
triplets $(e_s, r, e_o)$ equals the number of edges, \ie $|\mathcal{G}| = n$.

Formally, in a graph $G = (V, E)$, the \textbf{degree} of a vertex $v \in V$ is the number of
edges incident to it. When it comes to directed graphs, \textbf{in-degree} and
\textbf{out-degree} of a vertex $v$ is usually distinguished, which are defined as
\begin{align}
deg^{+}(v) & = |\{ (u, v)\ |\ u \in V, (u, v) \in E \}| \\
deg^{-}(v) & = |\{ (v, u)\ |\ u \in V, (v, u) \in E \}|
\end{align}

\noindent\begin{minipage}[h]{\columnwidth}
  \centering
  \captionsetup{format=customCaption,type=figure}
  \includegraphics[width=\columnwidth]{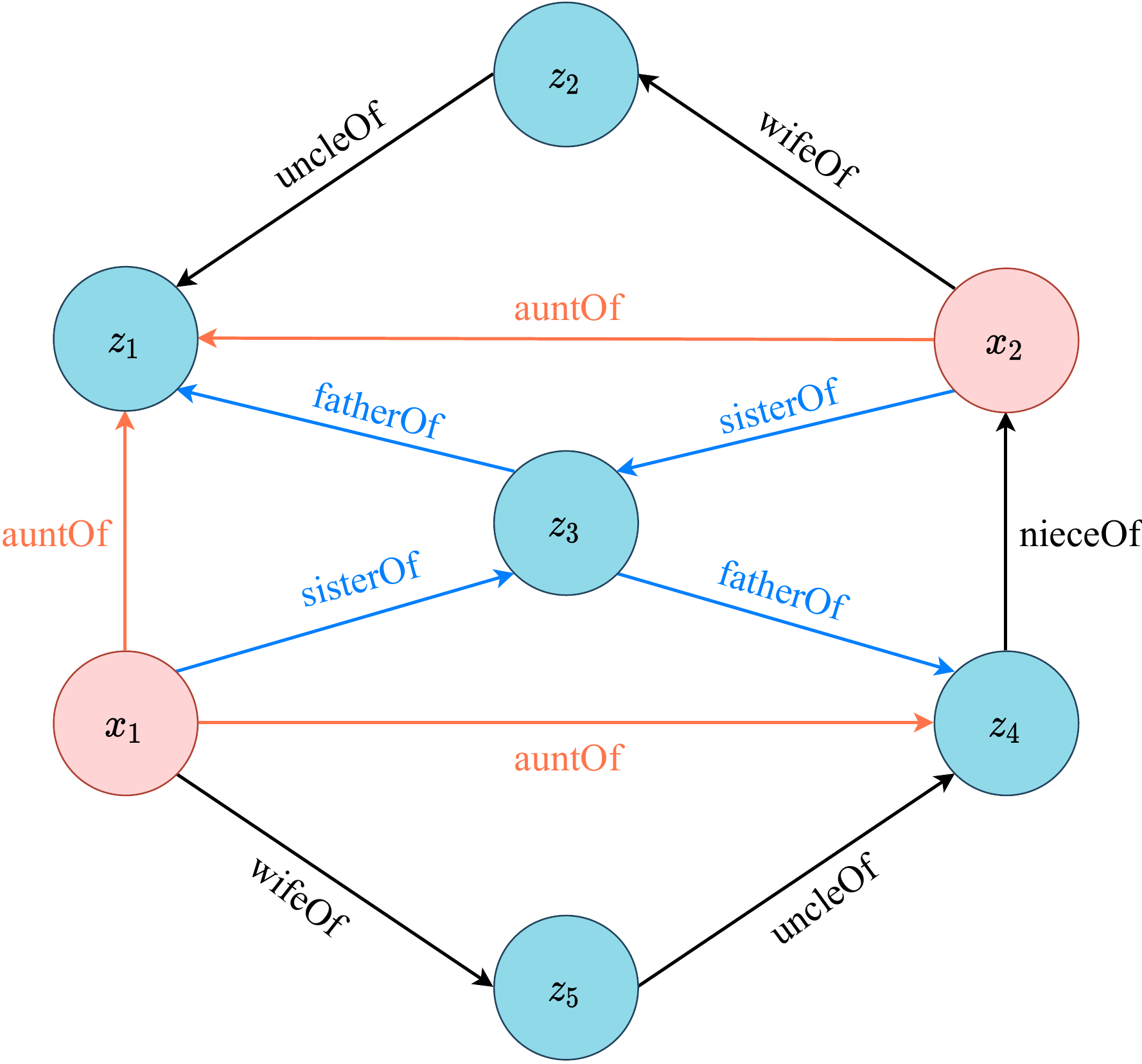} \\
  \caption{A KG example of family members and the relations between them.}
  \label{fig-matmul}
\end{minipage}

But in this paper, we use "degree" to represent the edges incident to a specific node $v$ for conciseness.

\begin{definition}[Heterogeneous Graph]\cite{wang2019heterogeneous}
A graph $G = (V, E)$ is heterogeneous when it consists a mapping function of node type $\phi: V \rightarrow \mathcal{A}$ and a mapping function of edge type $\psi: E \rightarrow \mathcal{R}$. $\mathcal{A}$ and $\mathcal{R}$ denote the sets of entity types and edge types (relations) in the corresponding KG.
\end{definition}

\marktext{
As shown in Fig. \ref{fig-matmul}, nodes of different types are marked in different colors, and edges are categorized by their relational labels.
}

\section{\blacktext{Methodology}} \label{sec-method}

\blacktext{
Capable of simultaneously learning representations and logical rules, Neural LP \cite{yang2017differentiable}
is the first differentiable neural system for knowledge graph reasoning that combines structure learning and
parameter learning. Our work follows the work of Neural LP and extensive studies based on it to consider
the problem of reasoning in heterogeneous KGs, from the view of mining intrinsic properties of the entities
in KGs.
}

\subsection{Neural LP for logic reasoning}
\subsubsection{TensorLog}\label{subsubsec:TensorLog}
The work of TensorLog \cite{cohen2016tensorlog, kathryn2018tensorlog} successfully simulates
the reasoning process using first-order logic rules by performing sparse matrix multiplication,
based on which, Neural LP \cite{yang2017differentiable} proposed a fully differentiable reasoning system.
In the following, we will first introduce the TensorLog operator.
In a KG involving a set of entities $\mathcal{E}$ and a set of relations $\mathcal{R}$,
factual triplets \textit{w.r.t.} the relation $r_k$ are restored in a binary matrix $\mathrm{M}_{r_k}$
$\in \{0, 1\}^{|\mathcal{E}| \times |\mathcal{E}|}$. $\mathrm{M}_{r_k}$, an adjacency matrix,
is called a TensorLog operator meaning that $(e_i, r_k, e_j)$ is in the KG if and only if
the $(i, j)$-th entry of $\mathrm{M}_{r_k}$ is 1. Let $\mathrm{v}_{e_i} \in \{0, 1\}^{|\mathcal{E}|}$
be the one-hot encoded vector of entity $e_i$,
then $s^{\top} = \mathrm{v}_{e_i}^{\top} \mathrm{M}_{r_1} \mathrm{M}_{r_2} \mathrm{M}_{r_3}$
is the \emph{path features vector} \cite{yang2019learn}, where the $j$-th entry counts the number of
unique rule paths following the pattern $r_1 \wedge r_2 \wedge r_3$ from $e_i$ to $e_j$ \cite{guu2015traversing}.

For example, every KG entity $e \in \mathcal{E}$ in Fig. \ref{fig-matmul} is encoded into a
$0 \mbox{-} 1$ vector of length $|\mathcal{E}| = 7$. For every relation $r \in \mathcal{R}$ and
every pair of entities $e_i, e_j \in \mathcal{E}$, the TensorLog operator relevant to $r$
is define as a sparse matrix $\mathrm{M}_r$ with its $(i, j)$-th element being 1
if $(e_i, r, e_j) \in \mathcal{G}$. Considering the KG in Fig. \ref{fig-matmul},
for the relation $r = \texttt{auntOf}$ we have

\begin{equation*}
  \mathrm{M}_r =
  \begin{blockarray}{cccccccc}
    x_1 & x_2 & z_1 & z_2 & z_3 & z_4 & z_5 \\
    \begin{block}{[ccccccc]c}
      0 & 0 & \textcolor{red}{1} & 0 & 0 & \textcolor{red}{1} & 0 & x_1 \\
      0 & 0 & \textcolor{red}{1} & 0 & 0 & 0 & 0 & x_2 \\
      0 & 0 & 0 & 0 & 0 & 0 & 0 & z_1 \\
      0 & 0 & 0 & 0 & 0 & 0 & 0 & z_2 \\
      0 & 0 & 0 & 0 & 0 & 0 & 0 & z_3 \\
      0 & 0 & 0 & 0 & 0 & 0 & 0 & z_4 \\
      0 & 0 & 0 & 0 & 0 & 0 & 0 & z_5 \\
    \end{block}
  \end{blockarray}
\end{equation*}

And the rule \texttt{sisterOf}(X, Z) $\wedge$ \texttt{fatherOf}(Z, Y) $\Rightarrow$ \texttt{auntOf}(X, Y)
can be simulated by performing the following sparse matrix multiplication:

\begin{equation*}
  \begin{gathered}
    \mathrm{M_{r'}} = \mathrm{M_{sisterOf}}\ \mathrm{M_{fatherOf}} = \\
    \begin{blockarray}{cccccccc}
      \begin{block}{[ccccccc]c}
        0 & 0 & \textcolor{red}{1} & 0 & 0 & \textcolor{red}{1} & 0 & x_1 \\
        0 & 0 & \textcolor{red}{1} & 0 & 0 & \textcolor{red}{1} & 0 & x_2 \\
        0 & 0 & 0 & 0 & 0 & 0 & 0 & z_1 \\
        0 & 0 & 0 & 0 & 0 & 0 & 0 & z_2 \\
        0 & 0 & 0 & 0 & 0 & 0 & 0 & z_3 \\
        0 & 0 & 0 & 0 & 0 & 0 & 0 & z_4 \\
        0 & 0 & 0 & 0 & 0 & 0 & 0 & z_5 \\
      \end{block}
      x_1 & x_2 & z_1 & z_2 & z_3 & z_4 & z_5 \\
    \end{blockarray}
  \end{gathered}
\end{equation*}
  
By setting $\mathrm{v}_{x_1} = [1, 0, 0, 0, 0, 0, 0]^{\top}$ as the one-hot vector of $x_1$ and
multiplying by $\mathrm{v}_{x_1}^{\top}$ on the left,
we obtain $s^{\top} = \mathrm{v}_{x_1}^{\top} \cdot \mathrm{M_{r'}} = [0, 0, 1, 0, 0, 1, 0]$.
The resultant $s^{\top}$ selects the row in $\mathrm{M_{r'}}$ actually identified by $x_1$.
By operating right-hand side multiplication with $\mathrm{v}_{z_1}$, we get the number of
unique paths following the pattern $\texttt{sisterOf} \wedge \texttt{fatherOf}$
from $x_1$ to $z_1$: $s^{\top} \cdot \mathrm{v}_{z_1} = 1$.

\subsubsection{Neural LP}\label{subsubsec-NeuralLP}

Neural LP \cite{yang2017differentiable} inherits the idea of TensorLog. Given a query
$q(h, t)$, after $L$ steps of reasoning, the score of the query induced through rule pattern
$\textbf{r}_s$ of length $L$ is computed as
\begin{equation}\label{eq:score}
\text{score}(t\ |\ q, h, \textbf{r}_s) =
\mathrm{v}^{\top}_h \prod_{l=1}^L \mathrm{M}^l \cdot \mathrm{v}_t,
\end{equation}
where $\mathrm{M}^l$ is the adjacency matrix of the relation used at the $l$-th hop.

The operators above are used to learn for query $q$ by calculating the weighted sum
of all possible patterns:
\begin{equation}\label{eq:weightedSum}
\sum_s{\alpha_s \prod_{k \in \beta_s}{\mathrm{M}_{r_k}}},
\end{equation}
where $s$ indexes over all potential patterns with maximum length of $L$,
$\alpha_s$ is the confidence score associated with the rule $\textbf{r}_s$
and $\beta_s$ is the ordered list of relations appearing in $\textbf{r}_s$.

To summarize, we update the score function in Eq. (\ref{eq:score})
by finding an appropriate $\alpha$ in
\begin{equation}\label{eq:varphi}
\varphi(t\ |\ q, h) =
\mathrm{v}^{\top}_h \sum_s{\alpha_s \cdot \left(\prod_{k \in \beta_s}{\mathrm{M}_{r_k}} \cdot \mathrm{v}_t \right)},
\end{equation}
and the optimization objective is
\begin{equation}\label{eq:objective}
\max_{\alpha_s}{\sum_{(h, q, t) \in \mathcal{G}} {\varphi(t\ |\ q, h)}},
\end{equation}
where $\alpha_s$ is to be learned.

Whereas the searching space of learnable parameters is exponentially large,
\ie $O(|\mathcal{R}|^L)$, direct optimization of Eq. (\ref{eq:objective})
may fall in the dilemma of over-parameterization. Besides,
it is difficult to apply gradient-based optimization. This is because each variable $\alpha_s$ is
bound with a specific rule pattern, and it is obviously a discrete work to enumerate rules.
To overcome these defects, the parameter of rule $\textbf{r}_s$ can be reformulated by distributing
the confidence to its inclusive relations at each hop, resulting in a differentiable score function:
\begin{equation}\label{eq:phi}
\phi_L(t\ |\ q, h) = \left(\mathrm{v}^{\top}_h \prod_{l=1}^{L}
{\sum_{k=0}^{|\mathcal{R}|}{a_k^l \mathrm{M}_{r_k}}} \right) \cdot \mathrm{v}_t,
\end{equation}
where $L$ is a hyperparameter denoting the maximum length of patterns and $|\mathcal{R}|$ is the
number of relations in KG. $\mathrm{M}_{r_0}$ is an identity matrix $I$ that enables the model
to include all possible rule patterns of length $L$ or smaller \cite{sadeghian2019drum}.

To perform training and prediction over the Neural LP framework, we should first construct a KG from
a large subset of all triplets. Then we remove the edge $(h, t)$ from the graph
when facing the query $(h, q, t)$, so that the score of $t$ can get rid of the influence imposed
from the head entity $h$ directly through the edge $(h, t)$ for the correctness of reasoning.

\subsection{\blacktext{Our DegreEmbed model}}\label{subsec:DegreEmbed}
In this section, we propose our \emph{DegreEmbed} model based on Neural LP \cite{yang2017differentiable}
as a combination of models relying on knowledge graph embedding and ILP models where the potential
properties of individual entities are considered through a technique we call degree embedding.
We discover that the attributes of nodes in a KG can make a difference via observation on their degrees.
In Fig. \ref{fig:heterogeneous}, we notice that Mike is a male because he is a nephew of someone,
hence it is incorrect indeed to reason by a rule containing a female-type relation starting from Mike.
Also, the in-degree (\ie \texttt{studyIn}) of entity THU proves its identity as a university.
Besides, as illustrated in Section \ref{subsubsec:TensorLog}, all knowledge of a KG
is stored in the relational matrices, which is our aim to reconstruct for harboring
type information of entities. For a query $q(h, t)$, the final score is a scalar obtained by 
Eq. (\ref{eq:phi}), where the \emph{path feature vector} is $s^{\top} = \mathrm{v}^{\top}_h
\prod_{l=1}^{L} {\sum_{k=0}^{|\mathcal{R}|}{a_k^l \mathrm{M}_{r_k}}}$, and $\mathrm{v}_t$ selects
the $t$-th element of $s^{\top}$ through matrix multiplication.
In fact, the vector $s^{\top} \in \mathbb{R}^{|\mathcal{E}|}$ is a row of matrix $\prod_{l=1}^{L}
{\sum_{k=0}^{|\mathcal{R}|}{a_k^l \mathrm{M}_{r_k}}}$, each value of which is 
the "influence" passed from head entity $h$ to the regarding entity.
As a result, we can consider the attributes of the entity $e_i$ by changing the $i$-th row of adjacency matrices
from the perspective of the type of degrees of $e_i$.

\begin{figure*}[t]
  \centering
  \includegraphics[width=\linewidth]{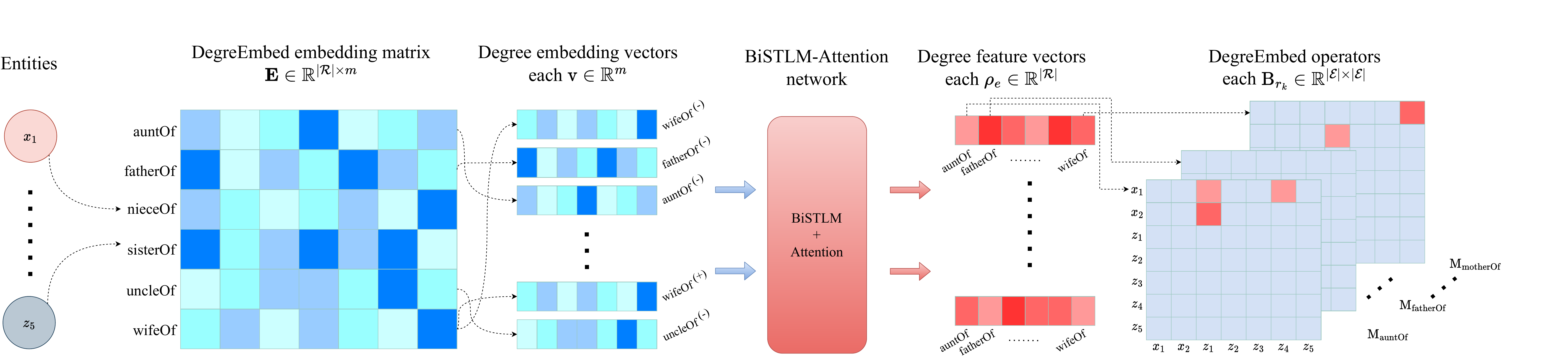}
  \caption{
    An illustration of computing the DegreEmbed operators for the KG shown in Fig. \ref{fig-matmul}.
    Superscripts (+) and (-) of the labels of degree embedding vectors denote their in and out direction.
    All DegreEmbed operators are initialized to zero matrices.
  }
  \label{fig:entity-embedding}
\end{figure*}

For any entity $e \in \mathcal{E}$, we collect the ones of unique types among all of its
in and out degrees separately to form a $d$-dimensional \emph{degree feature vector}, where
$d$ is the number of unique degrees. Then we project the vector onto a semantic space by
looking up in a row-vector embedding matrix $\textbf{E}^{|\mathcal{R}|\times m}$,
and the result is number of $d$ vectors arranged in a matrix $\textbf{M} \in \mathbb{R}^{d \times m}$,
where $m$ is the embedding dimension. The embedding vectors are input into a bidirectional LSTM
\cite{hochreiter1997long} at different time steps. Finally, we perform attention operation
on the hidden state of BiLSTM at the last time step to get the $|\mathcal{R}|$-dimensional
attention vector of $e$ for $1 \leq i \leq d$:
\begin{equation}
\textbf{h}_i, \textbf{h}'_{d-i+1} = \text{BiLSTM}(\textbf{h}_{i-1}, \textbf{h}'_{d-i}, \textbf{M}),
\label{eq-degree_embed}
\end{equation}
where $\textbf{h}$ and $\textbf{h}'$ are the hidden states of the forward and backward path LSTMs,
with the subscripts denoting their time step, and $H$, \textbf{the actual embedding vector} of entity $e$,
is obtained by concatenating $\textbf{h}_d$ and $\textbf{h}'_1$.

To compute the attention value on each relation imposed by entity $e$, we have
\begin{equation}
\rho_e = \text{softmax}(WH + b)
\end{equation}
Elements in $\rho_e \in \mathbb{R}^{|\mathcal{R}|}$ can be viewed as
the weights for relations. At last, we replace the elements that are in the row
identified by $e$ and equal 1 in each adjacency matrix $\mathrm{M}_{r_k}$ by
the $k$-th value of $\rho_e$. By following the same procedure for the other entities in the KG,
we construct a new set of relational matrices $B_{r_1}, \ldots, B_{r_{|\mathcal{R}|}}$,
which are called \emph{DegreEmbed} operators. The score function shown in Eq. (\ref{eq:phi}) is updated
accordingly as follows:
\begin{equation}\label{eq:phi'}
\phi'_L(t\ |\ q, h) = \left(\mathrm{v}^{\top}_h \prod_{l=1}^{L}
{\sum_{k=0}^{|\mathcal{R}|}{a_k^l \mathrm{B}_{r_k}}} \right) \cdot \mathrm{v}_t,  
\end{equation}
where the $\mathrm{B}_{r_1}, \ldots, \mathrm{B}_{r_{\mathcal{R}}}$ is our
new DegreEmbed operators, and $\mathrm{B}_{r_0}$ is still the identity matrix.
The whole process to compute the operators makes it possible to incorporate
the information of entities for rule learning models, where the degree feature vector $\rho_e$ can be
viewed as the identification of the entity $e$.
Remarkably, the DegreEmbed operators can be pre-trained due to its belonging to
the inner attribute of a KG, thus resulting in a model that can be easily deployed in similar tasks.
An overview of computing the DegreEmbed operators is illustrated in Fig. \ref{fig:entity-embedding}

Finally, the confidence scores are learned over the bidirectional LSTM
followed by the attention using Eqs (\ref{eq:lstm}) and (\ref{eq:attn}) for the temporal dependency
among several consecutive steps. The \texttt{input} in Eq. (\ref{eq:lstm})
is \texttt{query} embedding from another lookup table. For $1 \leq i \leq L$ we have
\begin{align}
& \textbf{h}_i, \textbf{h}'_{L-i+1} =
\text{BiLSTM}(\textbf{h}_{i-1}, \textbf{h}'_{L-i}, \text{input})\label{eq:lstm}, \\
& [a_{i, 1}, \ldots, a_{i, |\mathcal{R}|}] =
f_{\theta}\left( [\textbf{h}_i\ ||\ \textbf{h}'_{L-i}] \right), \label{eq:attn}
\end{align}
where $[a_{i, 1}, \ldots, a_{i, |\mathcal{R}|}]$ is the attention vector obtained
by performing a linear transformation over concatenated forward and
backward hidden states, followed by a softmax operator: $f_{\theta}(H) = \text{softmax} (WH + b)$.

\subsection{Optimization of the model}
\textbf{Loss construction}. In general, this task of link prediction is treated as a classification of entities
to build the loss. For each query $q(h, t)$ in a KG, we first split the objective function
Eq. (\ref{eq:phi'}) into two parts: target vector $\mathrm{v}_t$ and prediction vector
\begin{equation}
s^{\top} = \mathrm{v}^{\top}_h
\prod_{l=1}^{L} {\sum_{k=0}^{|\mathcal{R}|}{a_k^l \mathrm{B}_{r_k}}},
\end{equation}
and then our goal is to minimize the cross-entropy loss between $\mathrm{v}_t$ and $\mathrm{s}^{\top}$:
\begin{align*}
\ell_q(h, t) = -\sum_{i=1}^{|\mathcal{E}|} \{ & \mathrm{v}_t[i] \cdot \log{(\mathrm{s}[i])} + \\
& (1-\mathrm{v}_t[i]) \cdot \log{(1-\mathrm{s}[i])} \},
\end{align*}
where $i$ indexes elements in vector $\mathrm{v}_t$ and $\mathrm{s}$.

\textbf{Low-rank approximation}. It can be shown that the final confidences obtained by expanding
$\phi'_L$ are a rank one estimation of the \emph{confidence value tensor} \cite{sadeghian2019drum},
and low-rank approximation is a popular method for tensor approximation. Hence we follow the work
of \cite{sadeghian2019drum} and rewrite Eq. (\ref{eq:phi'}) using rank of $T$ approximation,
as shown in Eq. (\ref{eq:objective'}).
\begin{equation}\label{eq:objective'}
\varPhi_L(t\ |\ q, h) = \left(\mathrm{v}^{\top}_h \sum_{j=1}^T\prod_{l=1}^{L}
{\sum_{k=0}^{|\mathcal{R}|}{a_{j, k}^l \mathrm{B}_{r_k}}} \right) \cdot \mathrm{v}_t,  
\end{equation}

\begin{table*}[t]
  \caption{Statistics of datasets.}
  \label{tab:data-statistics}
  \centering
  \begin{tabular}{cccccccc}
    \toprule
    Dataset & \# Relation & \# Entity & \# Triplets & \# Facts & \# Train & \# Validation & \# Test \\
    \midrule
    FB15K-237 & 237 & 14541 & 310116 & 204087 & 68028 & 17535 & 20466 \\
    WN18 & 18 & 40943 & 151442 & 106088 & 35353 & 5000 & 5000 \\
    Family & 12 & 3007 & 28356 & 17615 & 5868 & 2038 & 2835 \\
    Kinship & 25 & 104 & 10686 & 6375 & 2112 & 1099 & 1100 \\
    UMLS & 46 & 135 & 6529 & 4006 & 3009 & 569 & 633 \\
    \bottomrule
  \end{tabular}
\end{table*}

More concretely, we update Eqs. (\ref{eq:lstm}) and (\ref{eq:attn}),
as is shown in Eqs. (\ref{eq:lstm'}) and (\ref{eq:attn'}),
by deploying number of $T$ BiLSTMs of the same network structure,
each of which can extract features from various dimensions.
\begin{align}
& \textbf{h}_i^{(j)}, \textbf{h}_{L-i+1}^{'(j)} =
\text{BiLSTM}_j(\textbf{h}_{i-1}^{(j)}, \textbf{h}_{L-i}^{'(j)}, \text{input})\label{eq:lstm'} \\
& [a_{i, 1}^{(j)}, \ldots, a_{i, |\mathcal{R}|}^{(j)}] =
f_{\theta}\left( [\textbf{h}_i^{'(j)}\ ||\ \textbf{h}_{L-i}^{'(j)}] \right), \label{eq:attn'}
\end{align}
where the superscripts of the hidden states identify their bidirectional LSTM.

\section{\blacktext{Experiment}} \label{sec-experiment}

In this section, we report the evaluation results of our model on a knowledge graph completion task,
where we compare the effectiveness of our model against the state-of-the-art learning systems for
link prediction. Meanwhile, as DegreEmbed takes advantage in the interpretability
in contrast to embedding-based methods, we also examine the rules mined by DegreEmbed with the help of
the indicator \emph{saturation}, which assesses the quality of rules from the corresponding
topological structure of a KG. We show that the top-scored rules mined by our method coincide with
those of high saturation scores, which in turn reflect the interpretability of our model.
To this end, we use ablation study to show how different components of our model contribute to its performance.

The knowledge graph completion task we use is a canonical one as described in \cite{yang2017differentiable}.
When training the model, the \texttt{query} and \texttt{head} are part of incomplete triplets
for training, and the goal is to find the most possible entity as the answer \texttt{tail}.
For example, if \texttt{nephewOf}(\texttt{Mike}, \texttt{Steve}) is missing from the
knowledge graph, our goal is to learn rules for reasoning over the existing KG
and retrieve \texttt{Steve} when presented with the query \texttt{nephewOf}(\texttt{Mike}, ?).

\subsection{Experiment setting}

\subsubsection{Datasets}\label{subsubsec:datasets}
To evaluate our method for learning logic rules in heterogeneous KGs, we select the following datasets
for knowledge graph completion task:

\begin{itemize}
  \item \emph{FB15K-237} \cite{toutanova2015observed}, a more challenging version of
    FB15K \cite{bordes2013translating} based on Freebase \cite{bollacker2008freebase},
    a growing knowledge graph of general facts.
  \item \emph{WN18} \cite{dettmers2018convolutional}, a subset of knowledge graph
    WordNet \cite{miller1995wordnet, miller1998wordnet} constructed for a widely used dictionary.
  \item \emph{Medical Language System (UMLS)} \cite{kok2007statistical}, from biomedicine,
    where the entities are biomedical concepts (\eg \texttt{organism}, \texttt{virus}) and
    the relations consist of \texttt{affects} and \texttt{analyzes}, etc.
  \item \emph{Kinship} \cite{kok2007statistical}, containing kinship relationships
    among members of a Central Australian native tribe.
  \item \emph{Family} \cite{kok2007statistical}, containing individuals from multiple families
    that are biologically related.
\end{itemize}

Statistics about each dataset used in our experiments are presented in Table \ref{tab:data-statistics}.
All datasets are randomly split into 4 files: \emph{facts}, \emph{train}, \emph{valid} and \emph{test},
and the ratio is 6:2:1:1. The \emph{facts} file contains a relatively large proportion of the triplets
for constructing the KG. The \emph{train} file is composed of query examples $q(h, t)$.
The \emph{valid} and \emph{test} files both contain queries $q(h, t)$,
in which the former is used for early stopping and the latter is for testing.
Unlike the case of learning embeddings, our method does not necessarily require the entities in
\emph{train}, \emph{valid} and \emph{test} to overlap.

\begin{table*}[t]
  \caption{
    Saturations of the Family dataset. $\gamma^{\textbf{p}_l}_q$, $\delta^{\textbf{p}_l}_q$,
    $\eta^{\textbf{p}_l}_q$ are \emph{macro}, \emph{micro} and \emph{comprehensive} saturations.
    The results relating to a specific relation are sorted by the comprehensive saturation in
    descending order.
  }
  \label{tab:saturation-family}
  \centering
  \begin{tabular}{rclccc}
    \toprule
    Rule & $\Rightarrow$ & Relation & $\gamma^{\textbf{p}_l}_q$ &
      $\delta^{\textbf{p}_l}_q$ & $\eta^{\textbf{p}_l}_q$ \\
    \midrule
    $X \xRightarrow{\texttt{brotherOf}} Z \xRightarrow{\texttt{brotherOf}} Y$ &
    $\Rightarrow$ & $X \xRightarrow{\texttt{brotherOf}} Y$ & .86 & .14 & .12 \\
    $X \xRightarrow{\texttt{nephewOf}} Z \xRightarrow{\texttt{uncleOf}} Y$ &
    $\Rightarrow$ &  & .77 & .13 & .10 \\
    $X \xRightarrow{\texttt{brotherOf}} Z \xRightarrow{\texttt{sisterOf}} Y$ &
    $\Rightarrow$ &  & .81 & .13 & .10 \\
    $X \xRightarrow{\texttt{sonOf}} Z \xRightarrow{\texttt{fatherOf}} Y$ &
    $\Rightarrow$ &  & 1.00 & .08 & .08 \\
    $X \xRightarrow{\texttt{nephewOf}} Z \xRightarrow{\texttt{auntOf}} Y$ &
    $\Rightarrow$ &  & .68 & .11 & .08 \\
    $X \xRightarrow{\texttt{sonOf}} Z \xRightarrow{\texttt{motherOf}} Y$ &
    $\Rightarrow$ &  & .98 & .07 & .07 \\
    \bottomrule
  \end{tabular}
\end{table*}

\subsubsection{\blacktext{Evaluation metrics}}
During training on the task of knowledge graph completion, for each triplet $(h, q, t)$,
two queries are designed as $(h, q, ?)$ and $(?, q, t)$ with answers
$t$ and $h$ for data augmentation. During evaluation, for each query, we manually remove
the edge $(h, t)$ from KG for the correctness of reasoning results and the score is computed
for each entity, as well as the rank of the correct answer. For the computed ranks from all
queries, we report the Mean Reciprocal Rank (MRR) and Hit@$k$ under the filtered protocol
\cite{bordes2013translating}. MRR averages the reciprocal ranks of the answer entities and
Hit@$k$ computes the percentage of how many desired entities are ranked among top $k$.

When evaluating the interpretability of rules, we choose a set of indicators called
\emph{macro}, \emph{micro} and \emph{comprehensive} saturations that measure the probability
of a rule pattern occurring in a certain relational subgraph $\mathcal{G}(r)$ from different angles.
More specifically, these computational methods analyze the reasoning complexity
from the inherent attributes of the graph structure $G$ \textit{w.r.t.} a KG $\mathcal{G}$.

\begin{definition}[Macro Reasoning Saturation]
Given a query $q \in \mathcal{R}$ and the maximum length $L$ of a rule pattern
$\textbf{r}_l \in \mathbb{H}^L$, the \textbf{macro reasoning saturation} of $\textbf{r}_l$
in relation to relation $q$, \ie $\gamma^{\textbf{r}_l}_q$, is the percentage of triples
$(h_i, q, t_j)$ in subgraph $\mathcal{G}(q)$ such that $\textbf{r}_l(h_i, t_j) \Rightarrow q(h_i, t_j)$.
\end{definition}

We can compute the macro reasoning saturation $\gamma^{\textbf{r}_l}_q$ using the following equation:
\begin{equation}\label{eq:gamma}
\gamma^{\textbf{r}_l}_q = \frac{|U^{\textbf{r}_l}|} {n^q},
\end{equation}
with $U^{\textbf{r}_l}$ being the set $U^{\textbf{r}_l} = \{ (h, q, t)\ |\ (h, q, t) \in
\mathcal{G}(q), \textbf{r}_l(h, t) \Rightarrow q(h, t) \}$ that collects the factual triplets
in $\mathcal{G}(q)$ as the reasoning candidates of rule $\textbf{r}_l$,
and $n^q = |\mathcal{G}(q)|$ being the number of edges (\ie the number of triples) in $G(q)$.
We can reasonably say that the larger $\gamma^{\textbf{r}_l}_q$ grows, the more likely
$\textbf{r}_l$ can be as a proper inference of the query $q$. When $\gamma^{\textbf{r}_l}_q$
equals 1, it means we can reason out every factual triple in $\mathcal{G}(q)$ through
at least one rule path following the pattern $\textbf{r}_l$.

\begin{definition}[Micro Reasoning Saturation]
Given the maximum length $L$ of a rule pattern, we define the
micro reasoning saturation of pattern $\textbf{r}_l \in \mathbb{H}^L$ as following.
Firstly, for a specific triple $\texttt{tri} = (h, q, t) \in \mathcal{G}$,
\ie $\delta^{\textbf{r}_l}_{\texttt{tri}}$, is the percentage of the paths
$\textbf{p}_{l_i} \rhd \textbf{r}_l$ such that $\textbf{r}_l(h, t) \Rightarrow q(h, t)$
as to all paths from $h$ to $t$.
\end{definition}

\begin{table*}[b]
  \caption{Knowledge graph completion performance comparison. Hit@$k$ (H@$k$) is in \%.}
  \label{tab:results}
  \centering
  \begin{tabular}{ccccccccccccc}
    \toprule
    & \multicolumn{4}{c}{Family} & \multicolumn{4}{c}{Kinship} & \multicolumn{4}{c}{UMLS} \\
    \cmidrule(lr){2-5} \cmidrule(lr){6-9} \cmidrule(lr){10-13}
    & MRR & H@1 & H@3 & H@10 & MRR & H@1 & H@3 & H@10 & MRR & H@1 & H@3 & H@10 \\
    \midrule
    TransE & .34 & 7 & 53 & 86 & .26 & 1 & 42 & 76 & .57 & 28 & 84 & 96 \\
    DistMult & .58 & 39 & 71 & 91 & .51 & 36 & 57 & 87 & .73 & 63 & 81 & 90 \\
    ComplEx & .83 & 72 & 94 & 98 & .61 & 44 & 71 & 92 & .79 & \textbf{69} & 87 & 95 \\
    TuckER & .43 & 28 & 52 & 72 & .60 & 46 & 70 & 86 & .73 & 63 & 81 & 91 \\
    RotatE & .90 & 85 & 95 & 99 & .65 & 50 & 76 & 93 & .73 & 64 & 82 & 94 \\
    \midrule
    RNNLogic & .93 & \textbf{91} & 95 & 99 & .64 & 50 & 73 & 93 & .75 & 63 & 83 & 92 \\
    Neural LP & .91 & 86 & 95 & 99 & .62 & 48 & 69 & 91 & .75 & 62 & 86 & 92 \\
    DRUM & .94 & 90 & 98 & 99 & .58 & 43 & 67 & 90 & \textbf{.80} & 66 & \textbf{94} & 97 \\
    DegreEmbed & \textbf{.95} & \textbf{91} & \textbf{99} & \textbf{100} &
      \textbf{.70} & \textbf{57} & \textbf{79} & \textbf{94} & \textbf{.80} &
      65 & \textbf{94} & \textbf{98} \\
    \bottomrule
  \end{tabular}
\end{table*}

The equation to compute $\delta^{\textbf{r}_l}_{\texttt{tri}}$ is
\begin{equation}
\delta^{\textbf{r}_l}_{\texttt{tri}} = \frac{|V^{\textbf{r}_l}|}{|V^{L}|}
\end{equation}
where $V^{\textbf{r}_l} = \{ \textbf{p}_{l_i}\ |\ \textbf{p}_{l_i} \rhd \textbf{r}_l,
\textbf{r}_l(h, t) \Rightarrow q(h, t) \}$,
$V^{L} = \{ \textbf{p}_{k_j}\ |\ \textbf{p}_{k_j} \rhd \textbf{r}_{k},
\forall \textbf{r}_{k} \in \mathbb{H}^L, \textbf{r}_{k}(h, t) \Rightarrow q(h, t) \}$.
$V^{\textbf{r}_l}$ denotes the set of rule paths derived from the pattern $\textbf{r}_l$
that is able to infer the fact $(h, q, t)$, and $V^{L}$ involves all the rules with their lengths
no longer that $L$.

Then, we average $\delta^{\textbf{r}_l}_{\texttt{tri}}$ over all triples $(h, q, t) \in \mathcal{G}(q)$
and get the \textbf{micro reasoning saturation} of the pattern $\textbf{r}_l \in \mathbb{H}^L$
for query $q$:
\begin{equation}\label{eq:delta}
\delta^{\textbf{r}_l}_q = \frac{1}{n^q}\sum_{\texttt{tri} \in \mathcal{G}(q)}
{\delta^{\textbf{r}_l}_{\texttt{tri}}}
\end{equation}

In Eqs. (\ref{eq:gamma}) and (\ref{eq:delta}),
$\gamma^{\textbf{r}_l}_q$ and $\delta^{\textbf{r}_l}_q$ assess how the probability
to infer $q$ following the pattern $\textbf{r}_l$ respectively from a macro and a micro perspective.
The higher the two indicators are, the easier for models to gain the inference
$\textbf{r}_l(h, t) \Rightarrow q(h, t)$. In order to obtain an overall result,
we define the comprehensive reasoning saturation $\eta^{\textbf{r}_l}_q$ by combining
the two indicators through multiplication, as revealed in Eq. (\ref{eq-comp_saturation}).
\begin{equation}\label{eq-comp_saturation}
\eta^{\textbf{r}_l}_q = \gamma^{\textbf{r}_l}_q \times \delta^{\textbf{r}_l}_q
\end{equation}
\blacktext{
We can imagine that the computation of comprehensive saturation on a certain logical rule $\textbf{r}_l$
to infer the relation $q$ involves two procedures: (1) select the triplets $(h, q, t)$ in subgraph $\mathcal{G}(q)$
that imply $\textbf{r}_l(h, t) \Rightarrow q(h, t)$ and (2) for each selected triplets,
calculate the percentage of rule paths following the pattern $\textbf{r}_l$ within all possible paths that
imply $q(h, t)$.
}

We can take the relation $q =$ \texttt{auntOf} and the rule
$\textbf{r}_l = $ \texttt{sisterOf} $\wedge$ \texttt{fatherOf}
in Fig. \ref{fig-matmul} as an example to show the computation of saturations.
In subgraph $\mathcal{G}(q)$, there are totally three triples (presented in red),
thus $n^q = 3$. For the triple $(x_2, \texttt{auntOf}, z_2)$, two rule paths can contribute to
its inference: \texttt{wifeOf}($x_2$, $z_1$) $\wedge$ \texttt{uncleOf}($z_2$, $z_1$)
and \texttt{sisterOf}($x_2$, $z_3$) $\wedge$ \texttt{fatherOf}($z_3$, $z_1$). In the same way, we can
see there are one and two rule paths for $(x_1, \texttt{auntOf}, z_1)$ and
$(x_1, \texttt{auntOf}, z_4)$ respectively. The rule $\textbf{r}_l = $ \texttt{sisterOf} $\wedge$ \texttt{fatherOf}
appears as an inference among all these three triples, therefore the macro saturation is
$\gamma^{\textbf{r}_l}_q = 3/n^q = 100\%$. More detailed information can be extracted through
computing the micro saturation. The rule $\textbf{r}_l$ takes a percentage of 50\% among all
paths for the triple $(x_2, \texttt{auntOf}, z_1)$, while 100\% and 50\% for the other two triples.
Thus, the micro saturation of $\textbf{r}_l$ for $q$ is $\delta^{\textbf{r}_l}_q = (0.5+1+0.5) / n^q = 67\%$.
Finally, we can compute the comprehensive saturation $\eta^{\textbf{r}_l}_q = \gamma^{\textbf{r}_l}_q
\times \delta^{\textbf{r}_l}_q = 67\%$.

We show a small subset of saturations computed from Family dataset in Table. \ref{tab:saturation-family}
for joint evaluation with logical rules mined by our model.
More results can be obtained in App. \ref{secapp-saturations}.

\subsubsection{Comparison of algorithms}\label{subsubsec:algorithms}
In experiment, the performance of our model is compared with that of the following algorithms:
\begin{itemize}
  \item Rule learning algorithms. Since our model is based on neural logic programming,
    we choose Neural LP and a Neural LP-based method DRUM \cite{sadeghian2019drum}.
    We also consider a probabilistic model called RNNLogic \cite{qu2020rnnlogic}.
  \item Embedding-based algorithms. We choose several embedding-based algorithms for comparison of
    the expressive power of our model, including TransE \cite{bordes2013translating},
    DistMult \cite{yang2014embedding}, ComplEx \cite{trouillon2016complex},
    TuckER \cite{balavzevic2019tucker} and RotatE \cite{sun2019rotate}.
\end{itemize}

The implementations of the above models we use are available at the links listed in App. \ref{secapp-urls}.

\begin{table*}[t]
  \caption{Top rules without reverse queries mined by DegreEmbed on the Family dataset.}
  \label{tab:rules}
  \centering
  \begin{tabular}{rclc}
    \toprule
    Rule & $\Rightarrow$ & Relation & Confidence \\
    \midrule
    $X \xRightarrow{\texttt{brotherOf}} Z \xRightarrow{\texttt{sisterOf}} Y$ &
    $\Rightarrow$ & $X \xRightarrow{\texttt{brotherOf}} Y$ & 1.00 \\
    $X \xRightarrow{\texttt{brotherOf}} Z \xRightarrow{\texttt{brotherOf}} Y$ &
    $\Rightarrow$ & & 0.81 \\
    $X \xRightarrow{\texttt{sonOf}} Z \xRightarrow{\texttt{motherOf}} Y$ &
    $\Rightarrow$ & & 0.55 \\
    $X \xRightarrow{\texttt{sonOf}} Z \xRightarrow{\texttt{fatherOf}} Y$ &
    $\Rightarrow$ & & 0.18 \\
    \bottomrule
  \end{tabular}
\end{table*}

\begin{table*}[t]
  \caption{Top rules without reverse queries mined by DRUM on the Family dataset.}
  \label{tab:rules-drum}
  \centering
  \begin{tabular}{rclc}
    \toprule
    Rule & $\Rightarrow$ & Relation & Confidence \\
    \midrule
    $X \xRightarrow{\texttt{sonOf}} Z \xRightarrow{\texttt{motherOf}} Y$ &
    $\Rightarrow$ & $X \xRightarrow{\texttt{brotherOf}} Y$ & 1.00 \\
    $X \xRightarrow{\texttt{brotherOf}} Z \xRightarrow{\texttt{brotherOf}} Y$ &
    $\Rightarrow$ & & 0.52 \\
    \textcolor{red}{$X \xRightarrow{\texttt{mohterOf}} Z \xRightarrow{\texttt{motherOf}} Y$} &
    $\Rightarrow$ & & 0.50 \\
    $X \xRightarrow{\texttt{sonOf}} Z \xRightarrow{\texttt{fatherOf}} Y$ &
    $\Rightarrow$ & & 0.48 \\
    $X \xRightarrow{\texttt{brotherOf}} Z \xRightarrow{\texttt{sisterOf}} Y$ &
    $\Rightarrow$ & & 0.35 \\
    \textcolor{red}{$X \xRightarrow{\texttt{motherOf}} Z \xRightarrow{\texttt{fatherOf}} Y$} &
    $\Rightarrow$ & & 0.13 \\
    \bottomrule
  \end{tabular}
\end{table*}

\subsubsection{Model configuration} \label{subsubsec-config}
Our model is implemented using PyTorch \cite{paszke2019pytorch} and the code will be publicly available.
We use the same hyperparameter setting during evaluation on all datasets. The query and entity embedding
have the dimension 128 and are both randomly initialized. The hidden state dimension of BiLSTM(s)
for entity and degree embedding are also 128.  As for optimization algorithm, we use mini-batch ADAM
\cite{kingma2014adam} with the batch size 128 and the learning rate initially set to 0.001.
We also observe that the whole model tends to be more trainable if we use the normalization skill.

Note that Neural LP \cite{yang2017differentiable}, DRUM \cite{sadeghian2019drum} and our
method all conform to a similar reasoning framework. Hence, to reach a fair comparison,
we ensure the same hyperparameter configuration during experiments on these models, where
the maximum rule length $L$ is 2 and the rank $T$ is 3 for DRUM and DegreEmbed, because Neural LP
is not developed using the low-rank approximation method.

\subsection{Results on KGC task}
We compare our DegreEmbed to several baseline models on the KGC benchmark datasets as stated in the
Section \ref{subsubsec:datasets} and Section \ref{subsubsec:algorithms}.
Our results on the selected benchmark datasets are summarized in Table \ref{tab:results} and
App. \ref{secapp-results}.

We notice that except that ComplEx \cite{trouillon2016complex} produces the best result among all
methods on UMLS under the evaluation of Hit@1, all models are outperformed by DegreEmbed with a
clear margin in Table \ref{tab:results}, especially on the dataset Kinship where we can see about
10\% improvement on some metrics. As expected, incorporating entity embedding enhances the expressive
power of DegreEmbed and thus benefits to reasoning on heterogeneous KGs.

In Table \ref{tab:results-more}, our model achieves state-of-the-art performance on WN18. It is
intriguing that embedding-based methods provide better predictions on FB15K-237 dataset,
with rule based methods, including RNNLogic, Neural LP, DRUM and ours, left behind.
As pointed in \cite{dettmers2018convolutional}, there are inverse relations from the training
data present in the test data in FB15K, which is called the problem of test set leakage,
resulting in the variant FB15K-237 where inverse relations are removed.
No wonder that methods depending on logic rule learning fails on this dataset.
\marktext{
Future work on more effective embedding representation of a node and its neighbor edges is likely to significantly advance the performance of link prediction models based on logic rules.
}

Notably, DegreEmbed not only is capable of producing state-of-the-art results on KGC task
thanks to the degree embedding of entities, but also maintains the advantage of logic rule
learning that enables our model to be interpretable to humans, which is of vital significance
in current research of intelligent systems. We will show the experiment results on the interpretability
of our DegreEmbed model later.

\subsection{\blacktext{Interpretability of our model}}

To demonstrate the interpretability of our method, we first report the logical rules mined by
our model and compare them with those by DRUM \cite{sadeghian2019drum}. Then we visualize the
embedding vectors learned through the proposed technique degree embedding to prove its expressive power.

\subsubsection{Quality of mined rules}
Apart from reaching state-of-the-art performance on KGC task which is largely thanks to the
mechanism of entity embedding, our DegreEmbed, as a knowledge graph reasoning model based on
logic rule mining, is of excellent interpretability as well. Our work follows the Neural LP
\cite{yang2017differentiable} framework, which successfully combines structure learning and
parameter learning to generate rules along with confidence scores.

In this section, we report evaluation results on explanations of our model where some of the rules
learned by DegreEmbed and DRUM are shown. As for evaluation metrics, we use the indicator \emph{saturations}
to objectively assess the quality of mined rules in a computable manner. We conduct two separate KGC experiments
for generating the logical rules where the only difference is whether the inverted queries are learned.
For better visualization purposes, experiments are done on the Family dataset, while other datasets
such as UMLS produce similar results.

We sort the rules by their normalized confidence scores, which are computed by dividing by
the maximum confidence of rules for each relation, and show top rules mined by our DegreEmbed and DRUM
without augmented queries respectively in Table \ref{tab:rules} and Table \ref{tab:rules-drum}.
Saturations of rules according to specific relations are shown in Table \ref{tab:saturation-family}.
For more results of saturations, learned rules \textit{w.r.t.} both pure and augmented queries,
please refer to the appendix.

By referring to the results given by computing saturations, we can see the rules mined by our model
solidly agree with the ones with high saturation level. Meanwhile, our model obviously gets rid of
the noises rendered by the heterogeneousness of the dataset through blending entity attributes
(\eg gender of entities) into rule learning. The rules mined for predicting the relation \texttt{brotherOf},
such as \texttt{brotherOf} $\wedge$ \texttt{sisterOf} and \texttt{brotherOf} $\wedge$ \texttt{brotherOf},
all show up with a male-type relation at the first hop.
However, there are logically incorrect rules mined by DRUM which are highlighted by \textcolor{red}{red}
in Table \ref{tab:rules-drum}. We think this is mainly because DRUM does not take entity attributes in to account.
In this case, our DegreEmbed model is capable of learning meaningful rules, which indeed proves
the interpretability of our model.

\noindent\begin{minipage}[h]{\columnwidth}
  \captionsetup{format=customCaption,type=figure}
  \centering
  \includegraphics[width=\columnwidth]{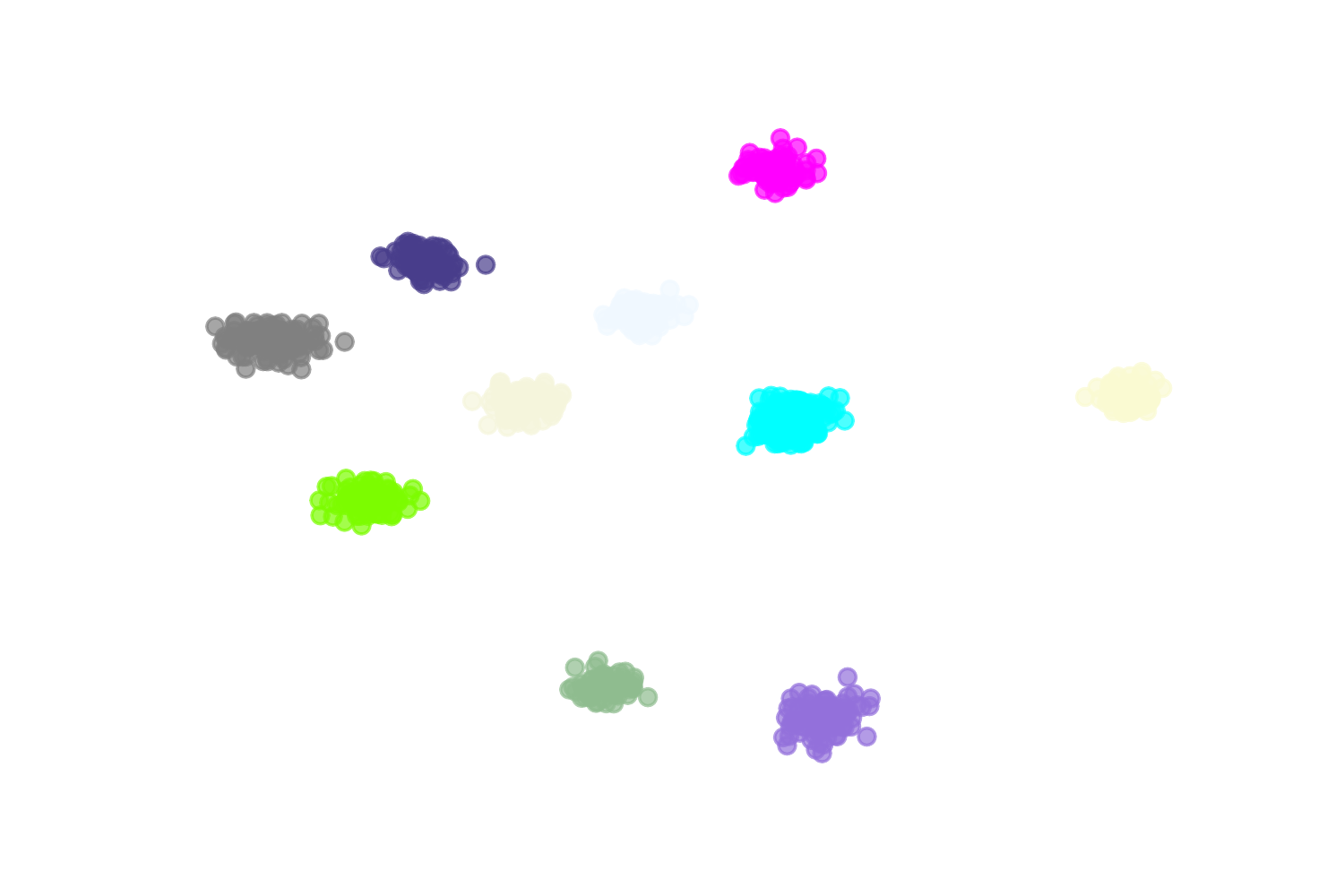}
  \caption{
    A t-SNE plot of the entity embedding of our trained model on Family dataset.
    Node colors denote their classes (\textit{i.e.} degree feature vectors).
  }
  \label{fig:embeddings_family}
\end{minipage}

\subsubsection{Learned entity embeddings}

To explain the learned degree embedding, we visualize the embeddings vectors of some entities from
the Family dataset. We use t-SNE \cite{van2008visualizing} to project the embeddings to two-dimensional
space and plot them in Fig. \ref{fig:embeddings_family}. In order to obtain the entity embeddings,
we first train our DegreEmbed model on Family with the same hyperparameter settings mentioned
in Section \ref{subsubsec-config}, and store the entire entity embedding matrix given by Eq. \ref{eq-degree_embed}.
Then, we classify the entities according to their \emph{degree feature vector} proposed
in Section \ref{subsec:DegreEmbed} and choose top ten most populated clusters marked with various colors
to plot in Fig. \ref{fig:embeddings_family}.

Note that, we use a logarithmic scale for the embedding plot to get better visualization results. In fact,
the representation of entities exhibits localized clustering in the projected 2D space, which verifies the
capability of our model to encode latent features of entities in heterogeneous KGs through their degrees.

\subsection{Ablation study}
\begin{figure*}[t]
  \centering
  \includegraphics[width=\linewidth]{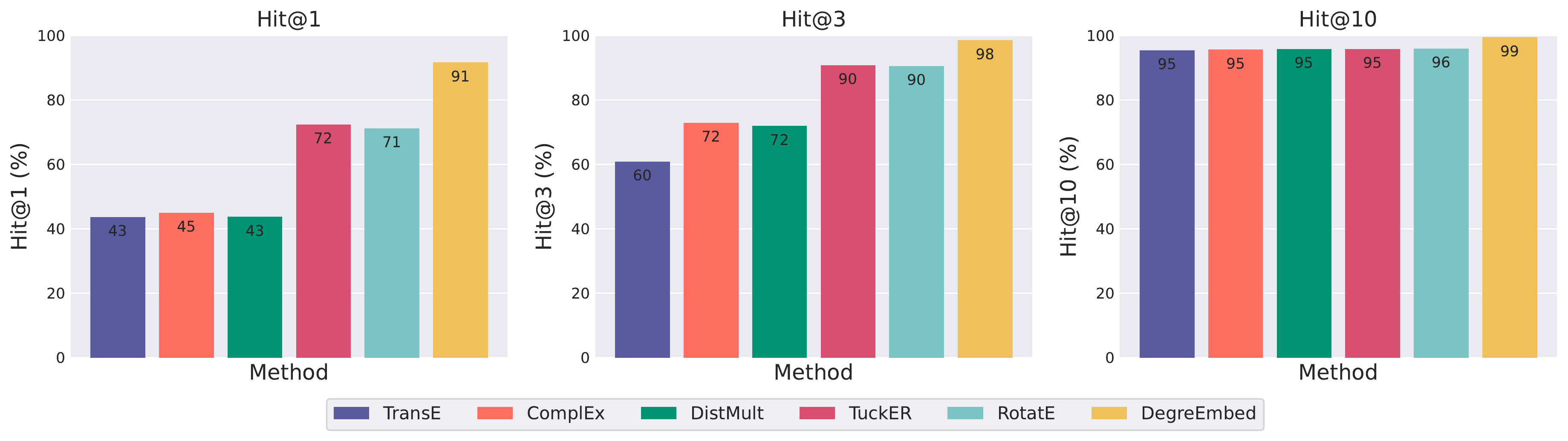}
  \caption{
    Model performance on Family with the original entity embeddings replaced by pre-trained ones
    from embedding-based methods. Hit@$k$ is in \%. The number inside each bar indicates
    its Hit@$k$ value.
  }
  \label{fig:bar_datasets}
\end{figure*}

\begin{figure*}[t]
  \centering
    \subfloat[Quantitative loss values showing the learning curve at training time.
              Various lengths of curves result from early stopping when model accuracy
              on validation set remains less or equal than the best within
              3 continuous training epochs.]
    {
      \includegraphics[width=\linewidth]{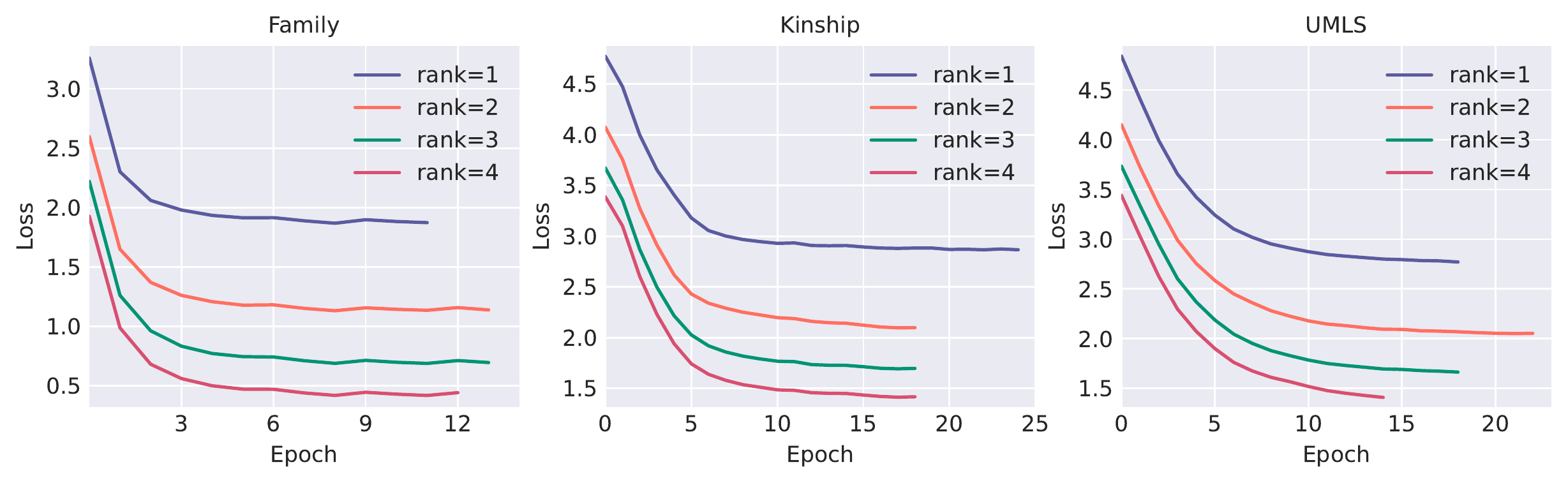}
      \label{fig:rank(a)}
    }
    \vspace{.01em}
    \subfloat[Model performance under the evaluation of Hit@$k$ on test datasets.]{
      \includegraphics[width=\linewidth]{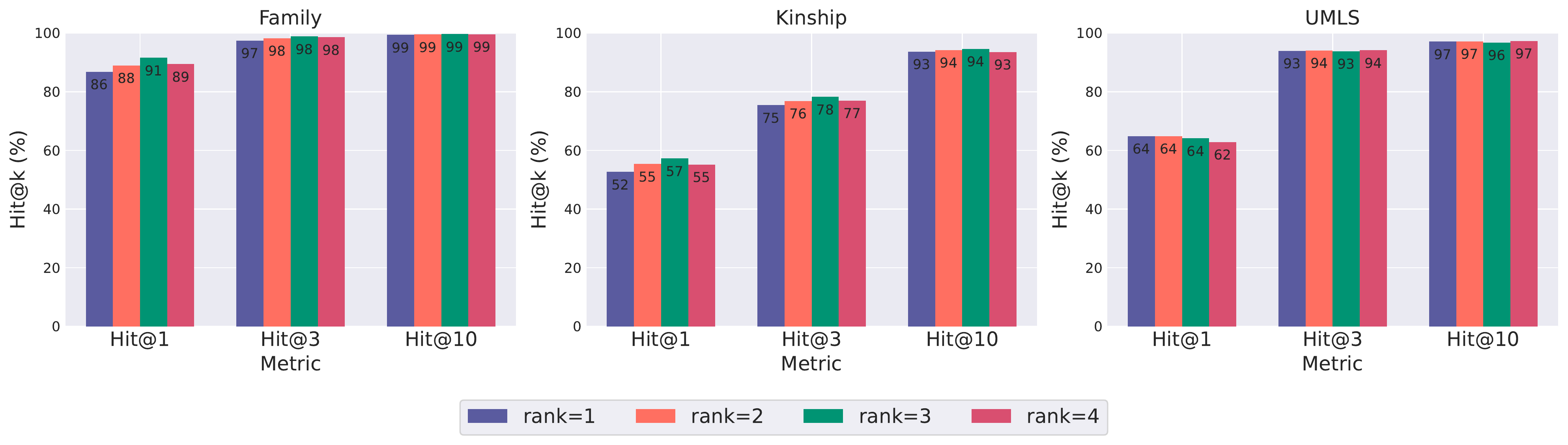}
      \label{fig:rank(b)}
    }
  \caption{Comparison among DegreEmbed variants with different ranks on three benchmark datasets.}
  \label{fig:rank}
\end{figure*}

To study the necessity of each component of our method, we gradually change the configuration
of each component and observe how the model performance varies.

\textbf{Degree embedding}. In this work, degree embedding is proposed as a new technique of entity embedding
for incorporating heterogeneous information in KGs. We successively replace this component with learned
entity embeddings from five pre-trained embedding-based models listed in Section \ref{subsubsec:algorithms}
on three datasets. We measure the Hit@1, Hit@3 and Hit@10 metrics and show the results on Family
in Fig. \ref{fig:bar_datasets}. Results on another two datasets are placed in the appendix.
In summary, the original model using degree embedding to encode entities produces the best results among all variants.
We hypothesis that this is due to the fact that many inner attributes of entities are lost in the embeddings
of those variants while DegreEmbed can learn to utilize these features implicitly.

\textbf{Low-rank approximation}. Tensor approximation of rank $T$ enables our model to learn latent features
from various dimensions, as show Eqs. (\ref{eq:lstm'}) and (\ref{eq:attn'}). We conduct experiments on three
datasets and show how model behavior differs with rank ranging from 1 to 4 in Fig. \ref{fig:rank}.
Training curves in Fig. (\ref{fig:rank(a)}) imply that model may converge faster with lower training loss
as rank goes up. However, Fig. \ref{fig:rank(b)} demonstrates that higher rank does not necessarily
bring better test results. We conjecture that this is because the amount of learnable features
of distinct dimensions varies from dataset to dataset, where the choice of rank matters a lot.
An intriguing insight can be obtained by combining Fig. (\ref{fig:rank(a)}) and Fig. (\ref{fig:rank(b)}):
training loss degrades as model rank increases while it barely contributes to results on test sets,
which provides a view of over-fitting.

\section{Conclusions} \label{sec-conclusions}
In this paper, a logic rule learning model called DegreEmbed has been proposed for reasoning
more effectively in heterogeneous knowledge graphs, where there exist entities and relations
of different types. Based on mining logic rules, DegreEmbed simultaneously leverages latent knowledge
of entities by learning embedding vectors for them, where the degrees of the entities are
closely observed. Experiment results show that our model benefits from the advantages of both
embedding-based methods and rule learning systems, as one can see DegreEmbed outperforms the
state-of-the-art models with a clear margin, and it produces high-quality rules with great interpretability.
In the future, we would like to optimize the way of entity embedding to increase the expressive
power of logic rule learning models for knowledge graph reasoning.

\begin{acks}
The work of this paper is supported by the "National Key R\&D Program of China" (2020YFB2009502),
"the Fundamental Research Funds for the Central Universities" (Grant No. HIT.NSRIF.2020098).
\end{acks}

\bibliographystyle{ios1}
\bibliography{bibliography}
\end{multicols}

\clearpage

\begin{appendix}

\section{Algorithm URLs} \label{secapp-urls}
\begin{table}[h]
  \caption{Available links to the models used in this work.}
  \label{tab:urls}
  \centering
  \begin{tabular}{lll}
    \toprule
    Algorithm & Link \\
    \midrule
    TransE, DistMult and ComplEx & \href{https://github.com/Accenture/AmpliGraph}{https://github.com/Accenture/AmpliGraph} \\
    TuckER & \href{https://github.com/ibalazevic/TuckER}{https://github.com/ibalazevic/TuckER} \\
    RotatE & \href{https://github.com/liyirui-git/KnowledgeGraphEmbedding\_RotatE}{https://github.com/liyirui-git/KnowledgeGraphEmbedding\_RotatE}  \\
    RNNLogic & \href{https://github.com/DeepGraphLearning/RNNLogic}{https://github.com/DeepGraphLearning/RNNLogic} \\
    Neural LP & \href{https://github.com/fanyangxyz/Neural-LP}{https://github.com/fanyangxyz/Neural-LP} \\
    DRUM & \href{https://github.com/alisadeghian/DRUM}{https://github.com/alisadeghian/DRUM} \\
    DegreEmbed (ours) & \href{https://github.com/lirt1231/DegreEmbed}{https://github.com/lirt1231/DegreEmbed} \\
    \bottomrule
  \end{tabular}
\end{table}

\section{Results on FB15K-237 and WN18} \label{secapp-results}
\begin{table}[h]
  \caption{
    Knowledge graph completion results on FB15K-237 and WN18. Hit@$k$ is in \%.
  }
  \label{tab:results-more}
  \centering
  \begin{tabular}{ccccccccc}
    \toprule
    & \multicolumn{4}{c}{FB15K-237} & \multicolumn{4}{c}{WN18} \\
    \cmidrule(lr){2-5} \cmidrule(lr){6-9}
    & MRR & Hit@1 & Hit@3 & Hit@10 & MRR & Hit@1 & Hit@3 & Hit@10 \\
    \midrule
    TransE & .15 & 5 & 19 & 25 & .36 & 4 & 63 & 81 \\
    DistMult & .25 & 17 & 28 & 42 & .71 & 56 & 83 & 93 \\
    ComplEx & .26 & 17 & 29 & 44 & .90 & 88 & 92 & 94 \\
    TuckER & \textbf{.36} & \textbf{27} & \textbf{39} & \textbf{46} & .94 & 93 & 94 & 95 \\
    RotatE & .34 & 24 & 38 & 53 & \textbf{.95} & \textbf{94} & \textbf{95} & 96 \\
    \midrule
    RNNLogic & .29 & 21 & 31 & 43 & .94 & 93 & 94 & 96 \\
    Neural LP & .25 & 19 & 27 & 37 & .94 & 93 & 94 & 95 \\
    DRUM & .25 & 19 & 28 & 38 & .54 & 49 & 54 & 66 \\
    DegreEmbed & .25 & 19 & 27 & 38 & \textbf{.95} & \textbf{94} & \textbf{95} & \textbf{97} \\
    \bottomrule
  \end{tabular}
\end{table}
\clearpage

\section{Extension to Table \ref{tab:saturation-family}: more saturations of Family} \label{secapp-saturations}
\begin{table}[h]
  \caption{
    Saturations of the Family dataset. $\gamma^{\textbf{p}_l}_q$, $\delta^{\textbf{p}_l}_q$,
    $\eta^{\textbf{p}_l}_q$ are \emph{macro}, \emph{micro} and \emph{comprehensive} saturations.
    The results relating to a specific relation are sorted by the comprehensive saturation in
    descending order.
  }
  \centering
  \begin{tabular}{rclccc}
    \toprule
    Rule & $\Rightarrow$ & Relation & $\gamma^{\textbf{p}_l}_q$ &
      $\delta^{\textbf{p}_l}_q$ & $\eta^{\textbf{p}_l}_q$ \\
    \midrule
    $X \xRightarrow{\texttt{nephewOf}} Z \xRightarrow{\texttt{brotherOf}} Y$ &
    $\Rightarrow$ & $X \xRightarrow{\texttt{nephewOf}} Y$ & .86 & .25 & .21 \\
    $X \xRightarrow{\texttt{nephewOf}} Z \xRightarrow{\texttt{sisterOf}} Y$ &
    $\Rightarrow$ &  & .79 & .22 & .17 \\
    $X \xRightarrow{\texttt{brotherOf}} Z \xRightarrow{\texttt{nephewOf}} Y$ &
    $\Rightarrow$ &  & .79 & .21 & .16 \\
    $X \xRightarrow{\texttt{brotherOf}} Z \xRightarrow{\texttt{nieceOf}} Y$ &
    $\Rightarrow$ &  & .72 & .17 & .12 \\
    $X \xRightarrow{\texttt{sonOf}} Z \xRightarrow{\texttt{brotherOf}} Y$ &
    $\Rightarrow$ &  & .64 & .10 & .06 \\
    $X \xRightarrow{\texttt{sonOf}} Z \xRightarrow{\texttt{sisterOf}} Y$ &
    $\Rightarrow$ &  & .36 & .05 & .02 \\
    \midrule
    $X \xRightarrow{\texttt{sisterOf}} Z \xRightarrow{\texttt{sonOf}} Y$ &
    $\Rightarrow$ & $X \xRightarrow{\texttt{daughterOf}} Y$ & .68 & .25 & .17 \\
    $X \xRightarrow{\texttt{sisterOf}} Z \xRightarrow{\texttt{daughterOf}} Y$ &
    $\Rightarrow$ &  & .61 & .20 & .12 \\
    $X \xRightarrow{\texttt{daughterOf}} Z \xRightarrow{\texttt{husbandOf}} Y$ &
    $\Rightarrow$ &  & .46 & .15 & .07 \\
    $X \xRightarrow{\texttt{daughterOf}} Z \xRightarrow{\texttt{wifeOf}} Y$ &
    $\Rightarrow$ &  & .46 & .14 & .06 \\
    \midrule
    $X \xRightarrow{\texttt{sisterOf}} Z \xRightarrow{\texttt{uncleOf}} Y$ &
    $\Rightarrow$ & $X \xRightarrow{\texttt{auntOf}} Y$ & .89 & .26 & .23 \\
    $X \xRightarrow{\texttt{sisterOf}} Z \xRightarrow{\texttt{auntOf}} Y$ &
    $\Rightarrow$ &  & .85 & .22 & .19 \\
    $X \xRightarrow{\texttt{auntOf}} Z \xRightarrow{\texttt{brotherOf}} Y$ &
    $\Rightarrow$ &  & .83 & .21 & .17 \\
    $X \xRightarrow{\texttt{auntOf}} Z \xRightarrow{\texttt{sisterOf}} Y$ &
    $\Rightarrow$ &  & .75 & .18 & .13 \\
    $X \xRightarrow{\texttt{sisterOf}} Z \xRightarrow{\texttt{fatherOf}} Y$ &
    $\Rightarrow$ &  & .66 & .09 & .06 \\
    $X \xRightarrow{\texttt{sisterOf}} Z \xRightarrow{\texttt{motherOf}} Y$ &
    $\Rightarrow$ &  & .34 & .05 & .02 \\
    \midrule
    $X \xRightarrow{\texttt{sisterOf}} Z \xRightarrow{\texttt{brotherOf}} Y$ &
    $\Rightarrow$ & $X \xRightarrow{\texttt{sisterOf}} Y$ & .89 & .15 & .13 \\
    $X \xRightarrow{\texttt{sisterOf}} Z \xRightarrow{\texttt{sisterOf}} Y$ &
    $\Rightarrow$ &  & .84 & .14 & .12 \\
    $X \xRightarrow{\texttt{nieceOf}} Z \xRightarrow{\texttt{uncleOf}} Y$ &
    $\Rightarrow$ &  & .78 & .13 & .10 \\
    $X \xRightarrow{\texttt{auntOf}} Z \xRightarrow{\texttt{nephewOf}} Y$ &
    $\Rightarrow$ &  & .67 & .12 & .08 \\
    $X \xRightarrow{\texttt{daughterOf}} Z \xRightarrow{\texttt{fatherOf}} Y$ &
    $\Rightarrow$ &  & 1.00 & .07 & .07 \\
    $X \xRightarrow{\texttt{daughterOf}} Z \xRightarrow{\texttt{motherOf}} Y$ &
    $\Rightarrow$ &  & .99 & .07 & .07 \\
    \midrule
    $X \xRightarrow{\texttt{brotherOf}} Z \xRightarrow{\texttt{sonOf}} Y$ &
    $\Rightarrow$ & $X \xRightarrow{\texttt{sonOf}} Y$ & .64 & .24 & .15 \\
    $X \xRightarrow{\texttt{brotherOf}} Z \xRightarrow{\texttt{daughterOf}} Y$ &
    $\Rightarrow$ &  & .56 & .19 & .10 \\
    $X \xRightarrow{\texttt{sonOf}} Z \xRightarrow{\texttt{husbandOf}} Y$ &
    $\Rightarrow$ &  & .46 & .16 & .08 \\
    $X \xRightarrow{\texttt{sonOf}} Z \xRightarrow{\texttt{wifeOf}} Y$ &
    $\Rightarrow$ &  & .46 & .14 & .06 \\
    $X \xRightarrow{\texttt{nephewOf}} Z \xRightarrow{\texttt{brotherOf}} Y$ &
    $\Rightarrow$ &  & .39 & .12 & .05 \\
    \bottomrule
  \end{tabular}
\end{table}
\clearpage

\section{Extension to Table \ref{tab:rules}: top rules obtained by our model}
\begin{table}[h]
  \caption{Top rules without reverse queries mined by DegreEmbed on the Family dataset.}
  \centering
  \begin{tabular}{rclc}
    \toprule
    Rule & $\Rightarrow$ & Relation & Confidence \\
    \midrule
    $X \xRightarrow{\texttt{brotherOf}} Z \xRightarrow{\texttt{nephewOf}} Y$ &
    $\Rightarrow$ & $X \xRightarrow{\texttt{nephewOf}} Y$ & 1.00 \\
    $X \xRightarrow{\texttt{brotherOf}} Z \xRightarrow{\texttt{nieceOf}} Y$ &
    $\Rightarrow$ & & 0.88 \\
    $X \xRightarrow{\texttt{sonOf}} Z \xRightarrow{\texttt{sisterOf}} Y$ &
    $\Rightarrow$ & & 0.34 \\
    $X \xRightarrow{\texttt{sonOf}} Z \xRightarrow{\texttt{brotherOf}} Y$ &
    $\Rightarrow$ & & 0.16 \\
    $X \xRightarrow{\texttt{nephewOf}} Z \xRightarrow{\texttt{sisterOf}} Y$ &
    $\Rightarrow$ & & 0.13 \\
    \midrule
    $X \xRightarrow{\texttt{sisterOf}} Z \xRightarrow{\texttt{sonOf}} Y$ &
    $\Rightarrow$ & $X \xRightarrow{\texttt{daughterOf}} Y$ & 1.00 \\
    $X \xRightarrow{\texttt{sisterOf}} Z \xRightarrow{\texttt{daughterOf}} Y$ &
    $\Rightarrow$ & & 0.84 \\
    $X \xRightarrow{\texttt{daughterOf}} Z \xRightarrow{\texttt{wifeOf}} Y$ &
    $\Rightarrow$ & & 0.72 \\
    $X \xRightarrow{\texttt{daughterOf}} Z \xRightarrow{\texttt{husbandOf}} Y$ &
    $\Rightarrow$ & & 0.24 \\
    \midrule
    $X \xRightarrow{\texttt{sisterOf}} Z \xRightarrow{\texttt{motherOf}} Y$ &
    $\Rightarrow$ & $X \xRightarrow{\texttt{auntOf}} Y$ & 1.00 \\
    $X \xRightarrow{\texttt{sisterOf}} Z \xRightarrow{\texttt{fatherOf}} Y$ &
    $\Rightarrow$ & & 0.77 \\
    $X \xRightarrow{\texttt{sisterOf}} Z \xRightarrow{\texttt{auntOf}} Y$ &
    $\Rightarrow$ & & 0.77 \\
    $X \xRightarrow{\texttt{sisterOf}} Z \xRightarrow{\texttt{uncleOf}} Y$ &
    $\Rightarrow$ & & 0.31 \\
    \midrule
    $X \xRightarrow{\texttt{sisterOf}} Z \xRightarrow{\texttt{sisterOf}} Y$ &
    $\Rightarrow$ & $X \xRightarrow{\texttt{sisterOf}} Y$ & 1.00 \\
    $X \xRightarrow{\texttt{sisterOf}} Z \xRightarrow{\texttt{brotherOf}} Y$ &
    $\Rightarrow$ & & 0.90 \\
    $X \xRightarrow{\texttt{sisterOf}} Z \xRightarrow{\texttt{motherOf}} Y$ &
    $\Rightarrow$ & & 0.39 \\
    \midrule
    $X \xRightarrow{\texttt{brotherOf}} Z \xRightarrow{\texttt{sonOf}} Y$ &
    $\Rightarrow$ & $X \xRightarrow{\texttt{sonOf}} Y$ & 1.00 \\
    $X \xRightarrow{\texttt{brotherOf}} Z \xRightarrow{\texttt{daughterOf}} Y$ &
    $\Rightarrow$ & & 0.67 \\
    $X \xRightarrow{\texttt{sonOf}} Z \xRightarrow{\texttt{husbandOf}} Y$ &
    $\Rightarrow$ & & 0.56 \\
    $X \xRightarrow{\texttt{sonOf}} Z \xRightarrow{\texttt{wifeOf}} Y$ &
    $\Rightarrow$ & & 0.39 \\
    \bottomrule
  \end{tabular}
\end{table}
\clearpage

\begin{table}[h]
  \caption{Top rules with reverse queries mined by DegreEmbed on the Family dataset.}
  \label{tab:rules-inv}
  \centering
  \begin{tabular}{rclc}
    \toprule
    Rule & $\Rightarrow$ & Relation & Confidence \\
    \midrule
    $X \xRightarrow{\texttt{inv\_sisterOf}} Z \xRightarrow{\texttt{inv\_uncleOf}} Y$ &
    $\Rightarrow$ & $X \xRightarrow{\texttt{nephewOf}} Y$ & 1.00 \\
    $X \xRightarrow{\texttt{brotherOf}} Z \xRightarrow{\texttt{inv\_auntOf}} Y$ &
    $\Rightarrow$ & & 0.39 \\
    $X \xRightarrow{\texttt{inv\_sisterOf}} Z \xRightarrow{\texttt{inv\_auntOf}} Y$ &
    $\Rightarrow$ & & 0.36 \\
    $X \xRightarrow{\texttt{inv\_brotherOf}} Z \xRightarrow{\texttt{inv\_auntOf}} Y$ &
    $\Rightarrow$ & & 0.32 \\
    $X \xRightarrow{\texttt{inv\_sisterOf}} Z \xRightarrow{\texttt{nieceOf}} Y$ &
    $\Rightarrow$ & & 0.29 \\
    $X \xRightarrow{\texttt{brotherOf}} Z \xRightarrow{\texttt{inv\_uncleOf}} Y$ &
    $\Rightarrow$ & & 0.22 \\
    \midrule
    $X \xRightarrow{\texttt{inv\_sisterOf}} Z \xRightarrow{\texttt{inv\_motherOf}} Y$ &
    $\Rightarrow$ & $X \xRightarrow{\texttt{daughterOf}} Y$ & 1.00 \\
    $X \xRightarrow{\texttt{inv\_sisterOf}} Z \xRightarrow{\texttt{inv\_fatherOf}} Y$ &
    $\Rightarrow$ & & 0.67 \\
    $X \xRightarrow{\texttt{inv\_brotherOf}} Z \xRightarrow{\texttt{inv\_fatherOf}} Y$ &
    $\Rightarrow$ & & 0.31 \\
    $X \xRightarrow{\texttt{sisterOf}} Z \xRightarrow{\texttt{inv\_fatherOf}} Y$ &
    $\Rightarrow$ & & 0.26 \\
    $X \xRightarrow{\texttt{inv\_brotherOf}} Z \xRightarrow{\texttt{inv\_motherOf}} Y$ &
    $\Rightarrow$ & & 0.18 \\
    $X \xRightarrow{\texttt{sisterOf}} Z \xRightarrow{\texttt{inv\_motherOf}} Y$ &
    $\Rightarrow$ & & 0.17 \\
    \midrule
    $X \xRightarrow{\texttt{brotherOf}} Z \xRightarrow{\texttt{inv\_sisterOf}} Y$ &
    $\Rightarrow$ & $X \xRightarrow{\texttt{brotherOf}} Y$ & 1.00 \\
    $X \xRightarrow{\texttt{inv\_brotherOf}} Z \xRightarrow{\texttt{inv\_brotherOf}} Y$ &
    $\Rightarrow$ & & 0.57 \\
    $X \xRightarrow{\texttt{brotherOf}} Z \xRightarrow{\texttt{brotherOf}} Y$ &
    $\Rightarrow$ & & 0.55 \\
    $X \xRightarrow{\texttt{brotherOf}} Z \xRightarrow{\texttt{sisterOf}} Y$ &
    $\Rightarrow$ & & 0.37 \\
    $X \xRightarrow{\texttt{inv\_brotherOf}} Z \xRightarrow{\texttt{inv\_sisterOf}} Y$ &
    $\Rightarrow$ & & 0.20 \\
    \midrule
    $X \xRightarrow{\texttt{inv\_sisterOf}} Z \xRightarrow{\texttt{motherOf}} Y$ &
    $\Rightarrow$ & $X \xRightarrow{\texttt{auntOf}} Y$ & 1.00 \\
    $X \xRightarrow{\texttt{sisterOf}} Z \xRightarrow{\texttt{fatherOf}} Y$ &
    $\Rightarrow$ & & 0.26 \\
    $X \xRightarrow{\texttt{inv\_sisterOf}} Z \xRightarrow{\texttt{inv\_nephewOf}} Y$ &
    $\Rightarrow$ & & 0.26 \\
    $X \xRightarrow{\texttt{inv\_sisterOf}} Z \xRightarrow{\texttt{inv\_nieceOf}} Y$ &
    $\Rightarrow$ & & 0.23 \\
    $X \xRightarrow{\texttt{inv\_sisterOf}} Z \xRightarrow{\texttt{inv\_daughterOf}} Y$ &
    $\Rightarrow$ & & 0.18 \\
    \midrule
    $X \xRightarrow{\texttt{sisterOf}} Z \xRightarrow{\texttt{sisterOf}} Y$ &
    $\Rightarrow$ & $X \xRightarrow{\texttt{sisterOf}} Y$ & 1.00 \\
    $X \xRightarrow{\texttt{sisterOf}} Z \xRightarrow{\texttt{inv\_brotherOf}} Y$ &
    $\Rightarrow$ & & 0.72 \\
    $X \xRightarrow{\texttt{inv\_sisterOf}} Z \xRightarrow{\texttt{inv\_sisterOf}} Y$ &
    $\Rightarrow$ & & 0.51 \\
    $X \xRightarrow{\texttt{inv\_brotherOf}} Z \xRightarrow{\texttt{inv\_sisterOf}} Y$ &
    $\Rightarrow$ & & 0.16 \\
    $X \xRightarrow{\texttt{inv\_sisterOf}} Z \xRightarrow{\texttt{inv\_brotherOf}} Y$ &
    $\Rightarrow$ & & 0.10 \\
    \midrule
    $X \xRightarrow{\texttt{inv\_brotherOf}} Z \xRightarrow{\texttt{inv\_motherOf}} Y$ &
    $\Rightarrow$ & $X \xRightarrow{\texttt{sonOf}} Y$ & 1.00 \\
    $X \xRightarrow{\texttt{inv\_sisterOf}} Z \xRightarrow{\texttt{inv\_fatherOf}} Y$ &
    $\Rightarrow$ & & 0.43 \\
    $X \xRightarrow{\texttt{brotherOf}} Z \xRightarrow{\texttt{inv\_motherOf}} Y$ &
    $\Rightarrow$ & & 0.37 \\
    $X \xRightarrow{\texttt{inv\_sisterOf}} Z \xRightarrow{\texttt{inv\_motherOf}} Y$ &
    $\Rightarrow$ & & 0.31 \\
    $X \xRightarrow{\texttt{brotherOf}} Z \xRightarrow{\texttt{inv\_fatherOf}} Y$ &
    $\Rightarrow$ & & 0.28 \\
    $X \xRightarrow{\texttt{inv\_brotherOf}} Z \xRightarrow{\texttt{inv\_fatherOf}} Y$ &
    $\Rightarrow$ & & 0.27 \\
    \bottomrule
  \end{tabular}
\end{table}
\clearpage

\section{Ablation results}
\begin{figure}[h]
  \centering
    \subfloat[Hit@1, Hit@3 and Hit@10 results on Kinship dataset. The number inside each bar indicates
              its Hit@$k$ value.]{
      \includegraphics[width=\linewidth]{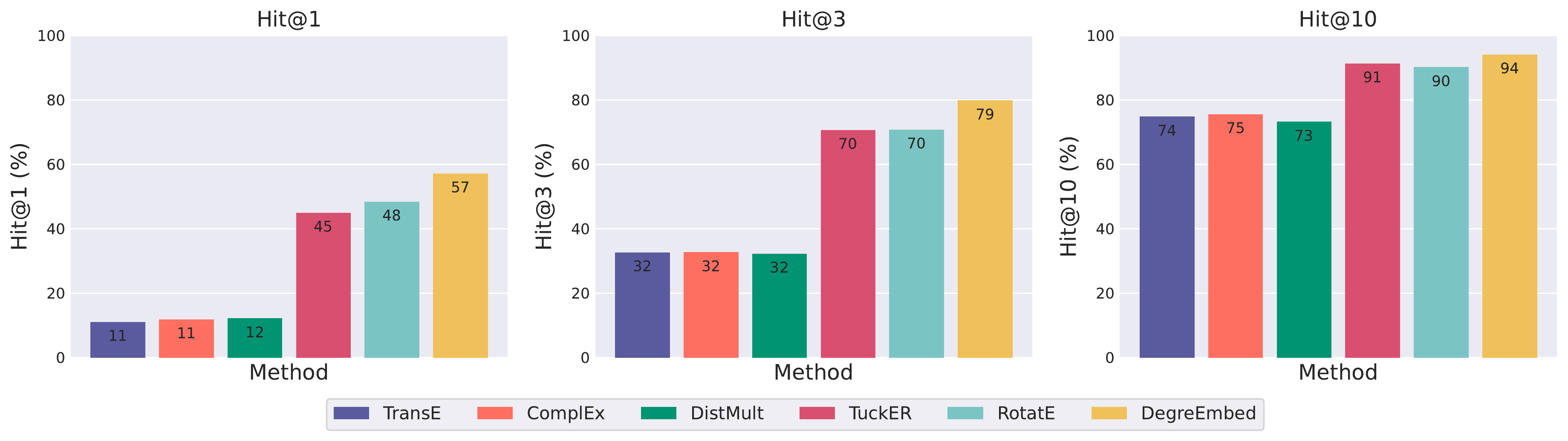}
      \label{fig:bar_datasets(b)}
    }
    \vspace{.01em}
    \subfloat[Hit@1, Hit@3 and Hit@10 results on UMLS dataset. The number inside each bar indicates
              its Hit@$k$ value.]{
      \includegraphics[width=\linewidth]{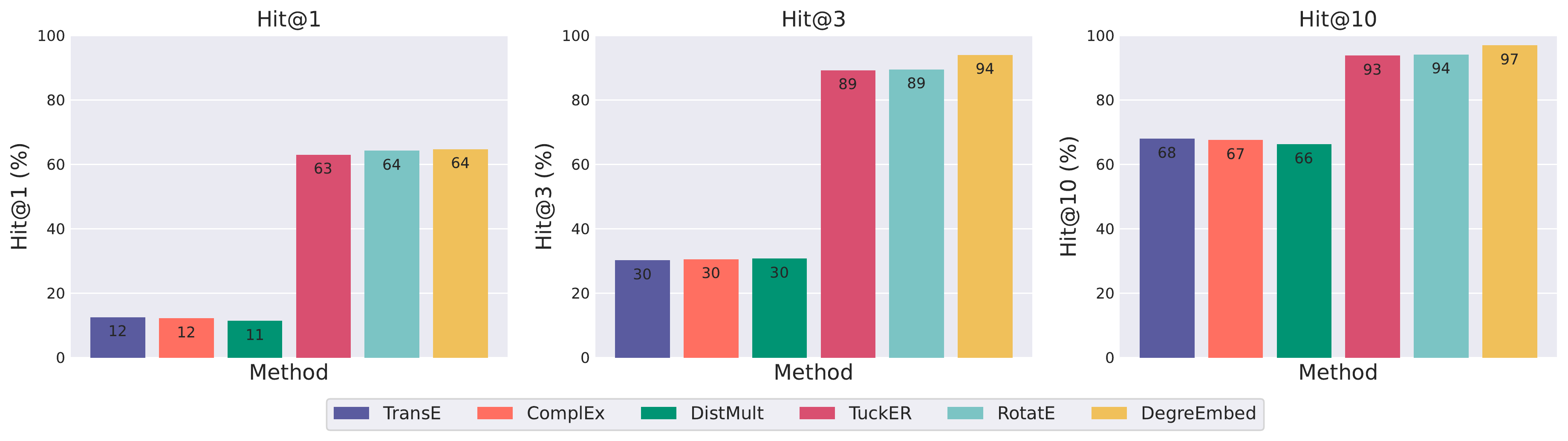}
      \label{fig:bar_datasets(c)}
    }
  \caption{
    Model performance on  Kinship and UMLS with the original entity embeddings replaced by
    pre-trained ones from embedding-based methods. Hit@$k$ is in \%.
  }
\end{figure}

\end{appendix}

\end{document}